\documentclass[journal]{IEEEtran}
\PassOptionsToPackage{bookmarks=false}{hyperref}
\usepackage{amsmath, amsthm, amsfonts, amssymb, mathrsfs, float}
\usepackage{algorithm}
\usepackage{algorithmic}
\usepackage{comment}
\usepackage{bm,bbm}
\usepackage{verbatim}
\usepackage{mathtools}
\usepackage{dsfont}
\usepackage{booktabs}
\usepackage{relsize}
\usepackage{caption}
\usepackage{subcaption}
\usepackage{color}
\usepackage[colorlinks,linkcolor=black,anchorcolor=black,citecolor=black,urlcolor=black]{hyperref}
\usepackage{cite}
\usepackage{multirow}
\theoremstyle{plain} 
\ifCLASSINFOpdf
\else
   \usepackage[dvips]{graphicx}
\fi
\usepackage{url}

\usepackage{graphicx}

\newtheorem{definition}{Definition}
\newtheorem{assumption}{Assumption}

\newtheorem{proposition}{Proposition}
\newtheorem{theorem}{Theorem}

\newcommand{\R}{\mathbb R}
\newcommand{\bbP}{\mathbb P}
\newcommand{\E}{\mathbb E}
\newcommand{\Q}{\mathbb Q}

\DeclareMathOperator*{\argmin}{argmin}
\DeclareMathOperator{\tr}{tr}

\begin{document}
\title{Distributionally Robust Graph Learning from Smooth Signals under Moment Uncertainty}

\author{Xiaolu Wang, Yuen-Man Pun, and Anthony Man-Cho So
\thanks{X. Wang, Y.-M. Pun and A. M.-C. So are with the Department of Systems Engineering and Engineering Management, The Chinese University of Hong Kong, Hong Kong SAR, China (e-mails: {\tt \{xlwang, ympun, manchoso\}@se.cuhk.edu.hk}). }}
\maketitle

\begin{abstract}
We consider the problem of learning a graph from a finite set of noisy graph signal observations, the goal of which is to find a smooth representation of the graph signal. Such a problem is motivated by the desire to infer relational structure in large datasets and has been extensively studied in recent years. Most existing approaches focus on learning a graph on which the observed signals are smooth. However, the learned graph is prone to overfitting, as it does not take the unobserved signals into account. To address this issue, we propose a novel graph learning model based on the distributionally robust optimization methodology, which aims to identify a graph that not only provides a smooth representation of but is also robust against uncertainties in the observed signals. On the statistics side, we establish out-of-sample performance guarantees for our proposed model. On the optimization side, we show that under a mild assumption on the graph signal distribution, our proposed model admits a smooth non-convex optimization formulation. We then develop a projected gradient method to tackle this formulation and establish its convergence guarantees. Our formulation provides a new perspective on regularization in the graph learning setting. Moreover, extensive numerical experiments on both synthetic and real-world data show that our model has comparable yet more robust performance across different populations of observed signals than existing non-robust models according to various metrics.
\end{abstract} 


\IEEEpeerreviewmaketitle

\vspace{-0.7\baselineskip}

\section{Introduction}
\subsection{Background and Motivation}
With the widespread availability of large, complex but structured datasets, one fundamental problem in contemporary data processing and analysis is that of inferring relationships among different entities using the data observed from them. Such a problem arises in many different application areas, including road traffic analysis, brain connectivity analysis, and community detection in social networks, just to name a few~\cite{ortega2018graph}. To tackle this problem, a common approach is to first model the entities as nodes of an undirected, weighted graph and the data observed from the entities as signals residing on the nodes, and then to learn the edges of the graph together with their weights based on the observed signals. Clearly, in order for the learning task to be well defined, it is necessary to specify how the signals are related to the graph topology. Towards that end, various models have been proposed in the literature; see, e.g.,~\cite{mateos2019connecting,dong2019learning} and the references therein. One representative model, which we shall refer to as the \emph{smooth graph signal} model and is motivated by considerations of real-world graph-structured data, postulates that the observed signals vary smoothly on the underlying graph---i.e., signal values at nodes that are adjacent to each other should be similar. In this model, a widely used measure of smoothness is the Laplacian quadratic form.
On one hand, the Laplacian quadratic form can be viewed as a discrete analog of the Dirichlet energy---a measure of variability for smooth functions---and has been used early on as a regularizer for learning problems on graphs~\cite{smola2003kernel,zhou2004regularization}. On the other hand, by drawing connections to classic signal processing concepts, one can interpret the eigenvectors of the Laplacian as frequency components and the eigenvalues as frequencies of the underlying graph~\cite{ortega2018graph,mateos2019connecting}. As such, the Laplacian quadratic form captures the variability of a given signal over the different graph frequency components. There has been a number of works that assume the smooth graph signal model and propose to learn the graph topology by solving a regularized Laplacian quadratic form minimization problem, where the regularizer is used to induce certain structure in the learned graph. For instance, the works~\cite{hu2013graph,dong2016learning} use a squared Frobenius norm regularizer to control the distribution of edge weights in the learned graph, while the work~\cite{kalofolias2016learn} combines the squared Frobenius norm with a logarithmic barrier to control both the sparsity and connectivity of the learned graph. Recently, some extensions of these formulations have been proposed, in which additional hard constraints are imposed on the graph topology; see, e.g.,~\cite{chepuri2017learning,kalofolias2018large,berger2020efficient}.

In order to assess the performance of a graph learning procedure, one possible approach is to first assume that the graph signal follows certain statistical model---for example, the Gaussian Markov random field (GMRF) or the factor analysis model (see, e.g.,~\cite{mateos2019connecting,dong2019learning} and the references therein)---and then evaluate the performance of the learned graph as a statistical estimator of the underlying graph. In the context of regularized Laplacian quadratic form minimization, if we assume that the graph signal is generated according to a ground-truth probability distribution (which in general is not known and depends on the underlying graph), then many existing formulations (such as those in~\cite{hu2013graph,dong2016learning,kalofolias2018large,berger2020efficient}) can be viewed as minimizing a regularized empirical risk of the observed signals, where the risk function is given precisely by the Laplacian quadratic form. Such a viewpoint raises the interesting possibility of analyzing the performance of these formulations using the vast array of tools developed in the statistical learning community for studying regularized empirical risk minimization (ERM) problems. Nevertheless, to the best of our knowledge, such a possibility has barely been pursued in the graph learning literature. In fact, two important questions concerning the regularized Laplacian quadratic form minimization approach remain unresolved:
\begin{itemize}
\item The approach essentially only uses the empirical distribution of the observed signals to learn the graph. As such, it can be prone to overfitting. In other words, the graph learned using the empirical distribution of the observed signals may differ greatly from the one learned if the ground-truth distribution of the graph signal were known. Is it possible to develop an alternative approach that can better exploit the information about the ground-truth distribution contained in the observed signals, so as to alleviate the effect of overfitting in the learned graph?

\item The regularizers used in existing formulations are usually ad-hoc in nature, and their impact on the quality of the learned graph is not well understood theoretically. Is there a more principled approach to regularization, so that one can construct regularizers whose effects on the learned graph can be rigorously explained?
\end{itemize}

\subsection{Our Contributions}


In this paper, we propose to take a distributionally robust optimization (DRO) approach to addressing the above questions. Specifically, with the risk function given by the Laplacian quadratic form, instead of minimizing the \emph{empirical risk} (i.e., expected risk with respect to (wrt) the empirical distribution) of the observed signals as in existing formulations, we minimize a \emph{worst-case expected risk}, where the worst-case expectation is taken wrt a set (called the \emph{ambiguity set}) of probability distributions that are consistent with certain information obtained from the observed signals. Intuitively, if the ambiguity set is chosen appropriately, then it contains the ground-truth distribution of the graph signal, which suggests that the graph learned by minimizing the worst-case expected risk not only provides a smooth representation of the graph signal but is also less susceptible to overfitting. Although there is a host of recent works that develop DRO-based techniques to tackle the issue of overfitting in statistical learning tasks (see, e.g.,~\cite{kuhn2019wasserstein}), our work is the first to pursue such an approach in the graph learning setting. Interestingly, our technical developments also lead to novel contributions to both the modeling and algorithmic aspects of DRO. Let us now summarize the main contributions of this paper.

\subsubsection{Modeling}
Using the fact that the risk function is quadratic in the graph signal, it is straightforward to show that for any given Laplacian, the expected risk depends only on the first two moments of the graph signal probability distribution. Based on this, we propose a novel moment-based distributionally robust graph learning model, in which the ambiguity set contains all distributions whose mean vectors and covariance matrices are close to the empirical mean and empirical covariance of the observed signals, respectively, and the goal is to find a Laplacian that minimizes the worst-case expected risk wrt such an ambiguity set. A notable feature of our proposed ambiguity set is that it depends on the decision variable of the model, namely the Laplacian. Such a dependence is crucial in the context of graph learning, as the probability distribution from which the observed signals are generated should depend on the underlying graph. However, the techniques currently available in the DRO literature for tackling decision-dependent ambiguity sets are very limited (see~\cite{luo2020distributionally} and the references therein) and do not apply to the setting considered in this work. This motivates us to develop new techniques to handle the ambiguity set in our proposed model; see ``\emph{Performance Analysis and Reformulation}'' below for further elaboration.

We remark that in view of the recent literature on DRO approaches to statistical learning, one may be tempted to consider a distributionally robust graph learning model similar to ours but with a \emph{Wasserstein distance}-based ambiguity set. Such an approach has in fact been explored in the recent work~\cite{zhang2021robust}, which appeared on arXiv at almost the same time as this work. However, the ambiguity set proposed in~\cite{zhang2021robust} does not depend on the Laplacian. Thus, the model in~\cite{zhang2021robust} fails to capture the interaction between the graph signal probability distribution and the structure of the underlying graph. In addition, since the expected risk does not distinguish between distributions with the same first and second moments, the Wasserstein distance-based ambiguity set carries much more information than is necessary for the purpose of evaluating the worst-case expected risk. This renders the subsequent analysis of the model more challenging and less direct than our moment-based model.

\subsubsection{Performance Analysis and Reformulation}
Under the assumption that the ground-truth distribution of the graph signal satisfies certain moment growth condition, we construct confidence regions around its mean and covariance by invoking the appropriate concentration inequalities. These regions not only provide a principled way of tuning the size of the ambiguity set but also yield a bound on the expected risk wrt the unknown ground-truth distribution of the graph signal (also known as the \emph{out-of-sample risk}). We then show that the worst-case expected risk minimization problem in our proposed model can be reformulated as a regularized ERM problem, in which the regularizers serve to promote robustness of the learned graph against the uncertainty described in the ambiguity set and the regularization parameters control the size of the ambiguity set. This establishes for the first time a rigorous link between distributional robustness and regularization in the graph learning setting.

\subsubsection{Algorithm Design and Analysis}
Although the aforementioned regularized ERM formulation has nice theoretical interpretations, it is computationally challenging to solve, as its objective function is generally non-smooth and non-convex. Nevertheless, we establish the curious result that as long as the ground-truth distribution of the graph signal has a probability density function, the objective function of the said formulation, though still non-convex, will be smooth almost surely. Consequently, we can apply a projected gradient descent (PGD) method to tackle the formulation. We show that the iterates generated by the PGD method converge to a stationary point of the regularized ERM formulation. Moreover, we prove that the same convergence result holds for various extensions of the formulation, which could be of independent interest. It is worth mentioning that our work contributes to the emerging area of algorithm design and analysis for DRO; for related works, see, e.g.,~\cite{li2019first,li2020fast}. To verify the efficacy of our proposed model and algorithm, we conduct extensive numerical experiments on both synthetic and real-world data. The results show that our proposed distributionally robust graph learning model is competitive---as measured by standard performance metrics---with several representative non-robust graph learning models in the literature. Moreover, when tested on different populations of observed signals, the former generally achieves a smaller variance in its performance than the latter. This demonstrates the value of incorporating distributional robustness in graph learning models.


\subsection{Notation}
The notation used in this paper is mostly standard. We use $\mathbb{S}^m$, $\mathbb{S}_+^m$, and $\mathbb{S}_{++}^m$ to denote the set of $m\times m$ symmetric, symmetric positive semidefinite, and symmetric positive definite matrices, respectively. We use $\bm{1}$ (resp.~$\bm{0}$) to denote the all-one (resp. all-zero) matrix, whose dimension will be clear from the context, and $\bm{I}_m$ to denote the $m\times m$ identity matrix. Given a matrix $\bm{A}\in\mathbb{S}^m$,  we use $A_{ij}$ to denote its $(i,j)$-th element, $\bm{A}^{\dagger}$ to denote its Moore--Penrose inverse (or pseudo-inverse), and $\|\bm{A}\|_F$ (resp. $\|\bm{A}\|$) to denote its Frobenius (resp.~spectral) norm. Given a probability distribution $\mathbb{Q}$, we write $\bm{x}\sim\mathbb{Q}$ to mean that the random vector $\bm{x}$ is distributed according to $\mathbb{Q}$ and $\mathbb{E}_{\bm{x}\sim\mathbb{Q}}[\cdot]$ to denote the expectation wrt $\mathbb{Q}$. We use $\Pr(\cdot)$ to denote probability, the distribution wrt which it is evaluated will be clear from the context. Given a vector $\bm{\mu}\in\mathbb{R}^m$ and a matrix $\bm{\Sigma} \in \mathbb{S}_+^m$, we use $\mathcal{N}(\bm{\mu},\bm{\Sigma})$ to denote the multivariate normal distribution with mean $\bm{\mu}$ and covariance $\bm{\Sigma}$.

\subsection{Paper Organization}
The rest of this paper is organized as follows: In Section \ref{sec-formulation}, we present our proposed moment-based distributionally robust graph learning model and study its statistical and optimization properties. Then, in Section~\ref{sec-pgd}, we discuss how the PGD method can be used to tackle our proposed model and analyze its convergence behavior under various settings. Next, we describe our experiment setups and report numerical results in Section \ref{sec-exp}. Finally, we close with some concluding remarks in Section \ref{sec-conclusion}.

\section{Distributionally Robust Graph Learning}\label{sec-formulation}
\subsection{Problem Formulation}\label{subsec:prob}
Consider $n$ given signals $\bm{x}^1,\ldots,\bm{x}^n \in \R^m$, where $x_i^j \in \R$ ($i=1,\ldots,m$; $j=1,\ldots,n$) denotes the $j$-th observed value at node $i$ of an unknown $m$-vertex, weighted, undirected graph $\mathcal{G}$. We view these signals as independent realizations of a random vector that follows a ground-truth distribution $\bbP^*$ associated with the graph $\mathcal{G}$; cf.~the statistical models for graph signals discussed in~\cite{dong2019learning} and the references therein. To identify the graph topology that yields a smooth representation of the graph signal, a popular strategy (see, e.g., \cite{hu2013graph,dong2016learning,kalofolias2016learn}) is to consider the following regularized Laplacian quadratic form minimization problem:
\begin{equation}\label{eq:gl-regular}
	\inf_{\bm{L}\in\mathcal{L}_s} \left\{ \frac{1}{n}\tr(\bm{X}^\top \bm{L} \bm{X}) + h(\bm{L}) \right\}.
\end{equation}
Here, $\bm{X} \coloneqq [ \bm{x}^1 \,\,\cdots\,\, \bm{x}^n] \in \R^{m\times n}$ is the data matrix whose columns are the observed signals, 
\[ 
	\mathcal{L}_s\coloneqq\left\{\bm{L} \in \mathbb{S}^m  : 
	\begin{aligned}
		&L_{ij} \leq 0 \text{\ for\ } i\neq j,\\
		&\bm{L}\bm{1}=\bm{0},\\
		&\tr(\bm{L})=2s
	\end{aligned}
	\right\}
\] 
is the set of $m\times m$ graph Laplacians whose scale is controlled by the parameter $s>0$, and $h:\mathbb{S}^m \rightarrow \R\cup\{+\infty\}$ is a convex regularizer that aims to promote certain structure in the learned Laplacian. As long as an optimal solution to Problem~\eqref{eq:gl-regular} exists and can be found efficiently, one can use it to construct the learned graph in a straightforward manner.

Although there have been extensive studies on the different choices of regularizer for Problem~\eqref{eq:gl-regular} and their effects on the learned graph, the ramifications of the fact that the graph learned by solving~\eqref{eq:gl-regular} depends on the particular realizations $\{\bm{x}^j\}_{j=1}^n$ are seldom addressed. To better understand the issue at hand, it is instructive to view Problem~\eqref{eq:gl-regular} through the lens of empirical risk minimization (ERM) in statistical learning. Specifically, let $\widehat{\bbP}_n$ be the empirical distribution associated with the data $\{\bm{x}^j\}_{j=1}^n$. Define $\mathcal{L}_s \times \R^n \ni (\bm{L},\bm{x}) \mapsto R({\bm L},\bm{x}) \coloneqq \bm{x}^\top \bm{L} \bm{x} \in \R_+$ to be the risk function. Since
\[ \E_{\bm{x} \sim \widehat{\bbP}_n}[ R(\bm{L},\bm{x}) ] = \frac{1}{n} \sum_{j=1}^n {\bm{x}^j}^\top \bm{L} \bm{x}^j  = \frac{1}{n} \tr(\bm{X}^\top \bm{L} \bm{X}), \]
we see that Problem~\eqref{eq:gl-regular} is equivalent to the following ERM problem:
\begin{equation} \label{eq:gl-erm}
\inf_{\bm{L}\in\mathcal{L}_s} \left\{ \E_{\bm{x} \sim \widehat{\bbP}_n}[ R(\bm{L},\bm{x}) ]  + h(\bm{L}) \right\}.
\end{equation}
Such a formulation reveals that an optimal solution $\widehat{\bm L}$ to Problem~\eqref{eq:gl-regular} may suffer from \emph{overfitting}---it yields a graph on which the \emph{observed} signals $\{\bm{x}^j\}_{j=1}^n$ are smooth but some \emph{unseen} signals generated according to $\bbP^*$ are not, so that the gap between the \emph{in-sample risk} $\E_{\bm{x} \sim \widehat{\bbP}_n}[ R(\widehat{\bm L},\bm{x}) ]$ and \emph{out-of-sample risk} $\E_{\bm{x} \sim \bbP^*}[ R(\widehat{\bm L},\bm{x}) ]$ is large. In principle, one can mitigate the effect of overfitting by choosing a suitable regularizer $h$. However, it is not easy to rigorously justify how a particular regularizer incorporates the information of the ground-truth distribution $\bbP^*$. Instead, we consider a DRO approach, in which we replace the \emph{empirical risk} $\E_{\bm{x} \sim \widehat{\bbP}_n}[ R(\bm{L},\bm{x}) ]$ in~\eqref{eq:gl-erm} by the \emph{worst-case expected risk}
\[ 
\sup_{\Q \in \mathcal{D}(\widehat{\bbP}_n)} \E_{\bm{x} \sim \Q}[ R(\bm{L},\bm{x}) ],
\] 
where $\mathcal{D}(\widehat{\bbP}_n)$, the so-called \emph{ambiguity set}, is a set of probability distributions that are ``close to'' $\widehat{\bbP}_n$. Intuitively, if the set $\mathcal{D}(\widehat{\bbP}_n)$ is small and contains $\bbP^*$, then the solution $\widetilde{\bm L}$ obtained by minimizing the worst-case expected risk over $\mathcal{L}_s$ will not be too conservative (i.e., it has a small worst-case expected risk) and will be less sensitive to the unseen signals (as $\widetilde{\bm L}$ would have taken the effect of $\bbP^*$ into account). To construct an ambiguity set with these desiderata, let us make the following simple yet crucial observation:
\begin{proposition}\label{prop:expectation}
Let $\Q$ be a probability distribution on the Borel $\sigma$-algebra in $\R^m$ with mean $\bm{\mu} \in \R^m$ and covariance $\bm{\Sigma} \in \mathbb{S}_+^m$; i.e., $\E_{\bm{x} \sim \Q}[\bm{x}] = \bm{\mu}$ and 
$\E_{\bm{x} \sim \Q}[ (\bm{x} - \bm{\mu})(\bm{x} - \bm{\mu})^\top ] = \bm{\Sigma}$. Then,
	\[
		\E_{\bm{x} \sim \Q} [ R(\bm{L},\bm{x}) ] = \tr( \bm{\Sigma} \bm{L} ) + \bm{\mu}^\top \bm{L} \bm{\mu}.
	\]
\end{proposition}
\begin{proof}
The result follows from a simple computation:
\begin{align*}
& \E_{\bm{x} \sim \Q} [ R(\bm{L},\bm{x}) ] = \E_{\bm{x} \sim \Q} [ \tr( \bm{x}\bm{x}^\top \bm{L} ) ] \\
&= \E_{\bm{x} \sim \Q} \left[ \tr\left( ( \bm{x} - \bm{\mu} )( \bm{x} - \bm{\mu})^\top \bm{L} + 2\bm{x}\bm{\mu}^\top\bm{L} - \bm{\mu}\bm{\mu}^\top\bm{L} \right) \right] \\
&= \tr(\bm{\Sigma}\bm{L}) + \bm{\mu}^\top \bm{L} \bm{\mu}.
\end{align*}
\end{proof}
\noindent Proposition~\ref{prop:expectation} shows that the expected risk $\E_{\bm{x} \sim \Q} [ R(\bm{L},\bm{x}) ]$ depends only on the mean and covariance of $\Q$. In particular, we have
\begin{equation} \label{eq:emp-smooth}
\E_{\bm{x} \sim \widehat{\bbP}_n} [ R(\bm{L},\bm{x}) ] = \tr( \widehat{\bm \Sigma}_n \bm{L} ) + \widehat{\bm\mu}_n^\top \bm{L} \widehat{\bm\mu}_n,
\end{equation}
where
\begin{equation} \label{eq:sample-stat}
\widehat{\bm\mu}_n \coloneqq \frac{1}{n} \sum_{j=1}^n \bm{x}^j, \,\,\, \widehat{\bm\Sigma}_n \coloneqq \frac{1}{n} \sum_{j=1}^n (\bm{x}^j-\widehat{\bm\mu}_n)(\bm{x}^j-\widehat{\bm\mu}_n)^\top
\end{equation}
are the empirical mean and empirical covariance of the observed signals, respectively. By the law of large numbers, we expect that as $n\rightarrow\infty$, the empirical mean $\widehat{\bm\mu}_n$ and the mean $\bm{\mu}^*$ of $\bbP^*$ will be close to each other, and the same is true for the empirical covariance $\widehat{\bm\Sigma}_n$ and the covariance $\bm{\Sigma}^*$ of $\bbP^*$. This suggests we should consider an ambiguity set that contains distributions whose mean vectors and covariance matrices are close to $\widehat{\bm\mu}_n$ and $\widehat{\bm\Sigma}_n$, respectively.

Concretely, let $\mathscr{P}(\bm{\mu},\bm{\Sigma})$ denote the set of probability distributions on the Borel $\sigma$-algebra in $\R^m$ with mean $\bm{\mu} \in \R^m$ and covariance $\bm{\Sigma} \in \mathbb{S}_+^m$. Given a Laplacian $\bm{L} \in \mathcal{L}_s$ and parameters $\rho_1,\rho_2>0$, we define the following ambiguity set:
\begin{align*} 
	 & \mathcal{M}(\bm{L},\rho_1,\rho_2) \\
	 & \coloneqq \left\{\Q \in \mathscr{P}(\bm{\mu},\bm{\Sigma}) : 
	\begin{aligned}
		&(\bm{\mu} - \widehat{\bm\mu}_n)^\top \bm{L} (\bm{\mu} - \widehat{\bm\mu}_n) \leq \rho_1^2, \\
		&\|\bm{\Sigma} - \widehat{\bm\Sigma}_n\|_F \leq \rho_2, \\
		&\bm{\mu} \in \R^m,\, \bm{\Sigma} \in \mathbb{S}_+^m
	\end{aligned}\right\}.
\end{align*}
In other words, every distribution in $\mathcal{M}(\bm{L},\rho_1,\rho_2)$ has its mean lying in the ellipsoid $\mathcal{E}(\widehat{\bm\mu}_n,\bm{L},\rho_1) \coloneqq \{ \bm{\mu} \in \R^m: (\bm{\mu} - \widehat{\bm\mu}_n)^\top \bm{L} (\bm{\mu} - \widehat{\bm\mu}_n) \leq \rho_1^2 \}$ and its covariance lying in the ball $\mathcal{B}(\widehat{\bm\Sigma}_n,\rho_2) \coloneqq \{ \bm{\Sigma} \in \mathbb{S}^m : \|\bm{\Sigma} - \widehat{\bm\Sigma}_n\|_F \leq \rho_2 \}$. On one hand, the use of a Frobenius-norm ball to describe a neighborhood of the empirical covariance $\widehat{\bm\Sigma}_n$ is rather intuitive. On the other hand, the use of an ellipsoid defined by $\bm{L}$ to describe a neighborhood of the empirical mean $\widehat{\bm\mu}_n$ is motivated by the factor analysis model proposed in~\cite{dong2016learning} for smooth graph signals. Indeed, suppose that the ground-truth Laplacian $\bm{L}^*$ admits the eigen-decomposition $\bm{L}^*=\bm{\chi} \bm{\Lambda} \bm{\chi}^\top$, where $\bm{\Lambda} = \text{Diag}(\lambda_1, \dots, \lambda_{m})$ is diagonal and $\bm{\chi} = [\bm{u}^1 \,\, \cdots \,\, \bm{u}^m ]$ is orthogonal. 
The factor analysis model in~\cite{dong2016learning} assumes that the graph signal $\bm{x}$ is generated as
\begin{equation}\label{eq-signal-gen}
	\bm{x} = \bm{\chi} \bm{r} + \bm{\mu}^* + \bm{\delta},	
\end{equation}
where $\bm{r} \sim \mathcal{N}(\bm{0},\bm{\Lambda}^\dagger)$ is the latent variable that controls the graph signal $\bm{x}$, $\bm{\mu}^* \in \R^m$ is the mean of $\bm{x}$, and $\bm{\delta} \sim \mathcal{N}(\bm{0},\epsilon^2\bm{I}_m)$ is the noise with power $\epsilon^2>0$. Under this model, we have $\bm{x} \sim \bbP^* = \mathcal{N}(\bm{\mu}^*, {\bm{L}^*}^\dagger + \epsilon^2\bm{I}_m)$; see~\cite{dong2016learning}. As such, it includes another widely-studied graph signal model in the literature, namely the GMRF model with a graph Laplacian precision matrix, as special case (see, e.g.,~\cite{egilmez2017graph} and the references therein). Since the observed signals $\{\bm{x}^j\}_{j=1}^n$ are assumed to be independent realizations of the random vector $\bm{x}$, we have $\widehat{\bm\mu}_n \sim \mathcal{N}\left( \bm{\mu}^*,\tfrac{1}{n}({\bm{L}^*}^\dagger + \epsilon^2\bm{I}_m) \right)$. Now, observe that for $i=1,\ldots,m$, the projection ${\bm{u}^i}^\top(\widehat{\bm\mu}_n-\bm{\mu}^*)$ of the deviation $\widehat{\bm\mu}_n-\bm{\mu}^*$ onto the $i$-th eigenbasis $\bm{u}^i$ of $\bm{L}^*$ is a mean-zero Gaussian random variable with variance
\begin{align*}
	\sigma_i^2 &= \E\left[\left( {\bm{u}^i}^\top (\widehat{\bm\mu}_n-\bm{\mu}^*) \right)^2 \right] \\ 
	&= \mathbb{E}\left[\tr\left( (\widehat{\bm\mu}_n-\bm{\mu}^*) (\widehat{\bm\mu}_n-\bm{\mu}^*)^\top \bm{u}^i {\bm{u}^i}^\top \right)\right] \\
	&= \frac{1}{n}\tr\left(\left( {\bm{L}^*}^\dagger+\epsilon^2 \bm{I}_m \right) \bm{u}^i {\bm{u}^i}^\top \right) = \frac{1}{n}(\nu_i+\epsilon^2)\|\bm{u}^i\|_2^2\\
	&= \frac{1}{n}(\nu_i+\epsilon^2),
\end{align*}
where $\nu_{i}=\tfrac{1}{\lambda_{i}}$ if $\lambda_{i}>0$ and $\nu_{i}=0$ if $\lambda_{i}=0$. In particular, if $\lambda_{i}>0$, then the larger the $\lambda_i$, the smaller the $\sigma_i^2$ and thus the more concentrated ${\bm{u}^i}^\top(\widehat{\bm\mu}_n-\bm{\mu}^*)$ is around $0$. Consequently, we expect that
\[ (\widehat{\bm\mu}_n-\bm{\mu}^*)^\top \bm{L}^* (\widehat{\bm\mu}_n-\bm{\mu}^*) = \sum_{i=1}^m \lambda_i \left({\bm{u}^i}^\top(\widehat{\bm\mu}_n-\bm{\mu}^*) \right)^2 \]
will be small with high probability (in fact, the above argument can be not only made rigorous but also extended to more general ground-truth distributions $\bbP^*$; see Theorem~\ref{thm-rho1}). This shows that given a Laplacian $\bm{L} \in \mathcal{L}_s$, we only need to take into account those distributions $\Q$ whose mean vectors $\bm{\mu}$ are close to the empirical mean $\widehat{\bm{\mu}}_n$ under the covariance structure induced by $\bm{L}$. Such a consideration gives rise to the ellipsoidal constraint $(\bm{\mu} - \widehat{\bm\mu}_n)^\top \bm{L} (\bm{\mu} - \widehat{\bm\mu}_n) \leq \rho_1^2$ in the definition of the ambiguity set $\mathcal{M}(\bm{L},\rho_1,\rho_2)$.

Based on the above development, we propose the following distributionally robust counterpart of the ERM problem~\eqref{eq:gl-erm}, which robustifies the learned graph against uncertainties about the ground-truth distribution $\bbP^*$:
\begin{equation}\label{eq:min-max}
\inf_{\bm{L} \in \mathcal{L}_s} \left\{  \sup_{\Q \in \mathcal{M}(\bm{L},\rho_1,\rho_2)} \E_{\bm{x} \sim \Q} [ R(\bm{L},\bm{x}) ] + h(\bm{L}) \right\}.
\end{equation}
Since we use the first moment $\widehat{\bm\mu}_n$ and second moment $\widehat{\bm\Sigma}_n$ to define the ambiguity set $\mathcal{M}(\bm{L},\rho_1,\rho_2)$, we shall refer to~\eqref{eq:min-max} as the \emph{Moment-Uncertain Graph Learning} (MUGL) model. Note that we keep the regularizer $h$ in the model, as it offers a way to induce structures beyond those that provide distributional robustness in the learned graph. A notable feature of the formulation~\eqref{eq:min-max} is that the ambiguity set $\mathcal{M}(\bm{L},\rho_1,\rho_2)$ \emph{depends} on the decision variable $\bm{L} \in \mathcal{L}_s$. This ensures that the distributions in $\mathcal{M}(\bm{L},\rho_1,\rho_2)$ reflect the graph structure encoded in $\bm{L}$. However, such a feature leads to reformulation and computational challenges that have not been addressed in the DRO literature before. As such, we need to develop new machinery to tackle our proposed MUGL model~\eqref{eq:min-max}.


\subsection{Bound on Out-of-Sample Risk}

Recall from our earlier discussion that it is desirable for an ambiguity set to be small and contain the ground-truth distribution $\bbP^*$. We now show that the ambiguity set $\mathcal{M}(\bm{L},\rho_1,\rho_2)$ will indeed possess such properties with high probability if the ground-truth distribution $\bbP^*$ satisfies certain moment growth condition and we choose the parameters $\rho_1,\rho_2$ judiciously. To begin, let us introduce the following definition:
\begin{definition}[Moment growth condition; cf.~\cite{so2011moment}] \label{def:mgc}
A probability distribution $\Q$ on the Borel $\sigma$-algebra in $\R^m$ with mean $\bm{\mu} \in \R^m$ is said to satisfy the \emph{moment growth condition} if there exists a constant $c>0$ such that for all $p\geq1$,
	\[
		\E_{\bm{x} \sim \Q} [\| \bm{x}-\bm{\mu} \|_2^p ] \leq (cp)^{p/2}.
	\]
\end{definition}

\noindent The moment growth condition defined above is rather mild. For instance, it is satisfied by any sub-Gaussian distribution~\cite{jin2019short}. In the remainder of this subsection, we make the following assumption:
\begin{assumption} \label{asp:cf-reg}
The ground-truth distribution $\bbP^*$ of the graph signal $\bm{x} \in \R^m$ has mean $\bm{\mu}^* \in \R^m$, covariance $\bm{\Sigma}^* \in \mathbb{S}_{+}^m$ and satisfies the moment growth condition. Let $\{\bm{x}^j\}_{j=1}^n$ be $n$ independent realizations of $\bm{x}$ and $\widehat{\bm\mu}_n, \widehat{\bm\Sigma}_n$ be given by~\eqref{eq:sample-stat}.
\end{assumption}

Now, using the probabilistic techniques developed in~\cite{so2011moment}, we can establish the following confidence region for the mean $\bm{\mu}^*$ of the ground-truth distribution $\bbP^*$:
\begin{theorem}[Confidence region for the mean] \label{thm-rho1}
	Suppose that Assumption~\ref{asp:cf-reg} holds. Let $\delta\in(0, e^{-2})$ be the confidence level. Then, there exists a constant $c_0>0$ such that for any $\bm{L} \in \mathcal{L}_s$, 
	\[ 
		(\widehat{\bm\mu}_n-\bm{\mu}^*)^\top \bm{L} (\widehat{\bm\mu}_n - \bm{\mu}^*) \leq \widehat{\rho}_1^2 \coloneqq \frac{4c_0e^2\ln^2(1/\delta)}{n}
	\] 
	will hold with probability at least $1-\delta$.
\end{theorem}
\begin{proof}
Since $\mathcal{L}_s$ is compact, we have $\max_{\bm{L} \in \mathcal{L}_s} \| \bm{L} \| < +\infty$. This, together with the assumption that $\bbP^*$ satisfies the moment growth condition, implies the existence of a constant $c_0>0$ such that for all $p\ge1$, 
\[ 
\max_{\bm{L} \in \mathcal{L}_s} \left\{ \E_{\bm{x} \sim \bbP^*} \left[ \| \bm{L}^{1/2} (\bm{x}-\bm{\mu}) \|_2^p \right] \right\} \leq (c_0p)^{p/2}.
\] 
The desired result then follows by adapting the proof of~\cite[Proposition 4]{so2011moment}.
\end{proof}

\noindent Moreover, we can establish the following confidence region for the covariance $\bm{\Sigma}^*$ of the ground-truth distribution $\bbP^*$:
\begin{theorem}[Confidence region for the covariance] \label{thm-rho2}
	Suppose that Assumption~\ref{asp:cf-reg} holds and $\bm{\Sigma}^* \in \mathbb{S}_{++}^m$. Let $\delta \in (0,e^{-2})$ be the confidence level. Then, there exist constants $c_1,c_2>0$ such that 
\begin{align*}
& \| \widehat{\bm\Sigma}_n - \bm{\Sigma}^* \|_F \le \widehat{\rho}_2 \\
&\coloneqq \frac{4c_1(2e/3)^{3/2}\ln^{3/2}(2m^{3/2}/\delta)}{n^{1/2}} \|\bm{\Sigma}^*\| + \frac{4c_2e^2\ln^2(2/\delta)}{n}
\end{align*}
will hold with probability at least $1-\delta$. 
\end{theorem}
\noindent The proof of the above result can be found in Appendix~\ref{app:cr-cov}. 

Theorems \ref{thm-rho1} and \ref{thm-rho2} imply that given a confidence level $\delta \in (0,e^{-2})$ and a Laplacian $\bm{L} \in \mathcal{L}_s$, the ambiguity set $\mathcal{M}(\bm{L},\widehat{\rho}_1,\widehat{\rho}_2)$ will contain the ground-truth distribution $\bbP^*$ with probability at least $1-2\delta$. Consequently, the out-of-sample risk bound
\[ \E_{\bm{x} \sim \bbP^*} [ R(\bm{L},\bm{x}) ] \le \sup_{\Q \in \mathcal{M}(\bm{L},\widehat{\rho}_1,\widehat{\rho}_2)} \E_{\bm{x} \sim \Q} [ R(\bm{L},\bm{x}) ] \]
will also hold with probability at least $1-2\delta$.


\subsection{Reformulation of the MUGL Model} \label{sec-reform}

Since $\mathcal{M}(\bm{L},\rho_1,\rho_2)$ is a set of probability distributions, it may seem at first sight that the inner supremum in the MUGL model~\eqref{eq:min-max} is an infinite-dimensional optimization problem. Nevertheless, using Proposition~\ref{prop:expectation} and the definition of $\mathcal{M}(\bm{L},\rho_1,\rho_2)$, we can reformulate~\eqref{eq:min-max} as the following finite-dimensional optimization problem:
\begin{equation}\label{eq:min-max-equiv}
	\begin{split}
		\inf_{\bm{L}} & \,\, \left\{ \sup_{\bm{\mu},\bm{\Sigma}}  \left\{ \tr(\bm{\Sigma} \bm{L}) + \bm{\mu}^\top \bm{L} \bm{\mu} \right\} + h(\bm{L}) \right\} \\
		\mbox{s.t.} & \,\,\, (\bm{\mu} - \widehat{\bm\mu}_n)^\top \bm{L} (\bm{\mu} - \widehat{\bm\mu}_n) \leq \rho_1^2, \\
		&\,\,\, \|\bm{\Sigma} - \widehat{\bm\Sigma}_n\|_F \leq \rho_2, \\
		&\,\,\, \bm{L} \in \mathcal{L}_s, \, \bm{\mu} \in \R^m,\, \bm{\Sigma} \in \mathbb{S}_+^m.
	\end{split}
\end{equation}
Observe that for any given $\bm{L} \in \mathcal{L}_s$, the inner supremum in~\eqref{eq:min-max-equiv} is separable in the variables $\bm{\mu} \in \R^m$ and $\bm{\Sigma} \in \mathbb{S}_+^m$. Hence, we can express Problem~\eqref{eq:min-max-equiv} as
\[ \inf_{\bm{L} \in \mathcal{L}_s} \{ \varphi_1(\bm{L}) + \varphi_2(\bm{L}) + h(\bm{L}) \}, \]
where
\begin{equation}\label{eq:max-mu}
\begin{split}
	\varphi_1(\bm{L}) \coloneqq& \ \sup_{\bm{\mu} \in \R^m} \bm{\mu}^\top \bm{L} \bm{\mu} \\
	&\ \ \ \text{s.t.}\ \ (\bm{\mu} - \widehat{\bm\mu}_n)^\top \bm{L} (\bm{\mu} - \widehat{\bm\mu}_n) \leq \rho_1^2
\end{split}
\end{equation}
and
\begin{equation}\label{eq:max-sigma}
	\begin{split}
		\varphi_2(\bm{L}) \coloneqq& \ \sup_{\bm{\Sigma} \in \mathbb{S}_+^m}  \tr ( \bm{\Sigma} \bm{L} ) \\
		& \ \ \ \text{s.t.} \ \ \|\bm{\Sigma}-\widehat{\bm\Sigma}_n\|_F \leq \rho_2.
	\end{split}
\end{equation}
As it turns out, both optimal value functions $\varphi_1$ and $\varphi_2$ have closed-form expressions.
\begin{proposition} \label{prop:closed-form}
For any given $\bm{L} \in \mathcal{L}_s$, the optimal values of Problems~\eqref{eq:max-mu} and~\eqref{eq:max-sigma} are given by
\begin{align*}
\varphi_1(\bm{L}) &= \left( \| \bm{L}^{1/2} \widehat{\bm\mu}_n \|_2 + \rho_1 \right)^2, \\
\varphi_2(\bm{L}) &= \tr(\widehat{\bm\Sigma}_n \bm{L}) + \rho_2 \|\bm{L}\|_F,
\end{align*}
respectively.
\end{proposition}
\noindent The proof of Proposition~\ref{prop:closed-form} can be found in Appendix~\ref{app:closed-form}. 

Using Proposition~\ref{prop:closed-form} and the identity~\eqref{eq:emp-smooth}, we obtain the following reformulation of the MUGL model~\eqref{eq:min-max}:
\begin{equation}\label{eq:reform}
\inf_{\bm{L}\in\mathcal{L}_s} \left\{ \frac{1}{n} \tr(\bm{X}^\top \bm{L} \bm{X}) + 2\rho_1 \| \bm{L}^{1/2} \widehat{\bm\mu}_n \|_2 + \rho_2\|\bm{L}\|_F + h(\bm{L}) \right\}.
\end{equation}
Compared with the non-robust graph learning model~\eqref{eq:gl-regular}, the distributionally robust MUGL model~\eqref{eq:reform} has two additional regularizers $h_1(\bm{L}) \coloneqq 2\rho_1 \| \bm{L}^{1/2} \widehat{\bm\mu}_n \|_2$ and $h_2(\bm{L}) \coloneqq \rho_2\|\bm{L}\|_F$. The regularizer $h_1$ (resp.~$h_2$) can be understood as promoting robustness of the learned graph against uncertainty about the mean (resp. covariance) of the ground-truth distribution $\bbP^*$, with the size $\rho_1$ (resp.~$\rho_2$) of the uncertainty region around the empirical mean $\widehat{\bm\mu}_n$ (resp. empirical covariance~$\widehat{\bm\Sigma}_n$) serving as the regularization parameter. Although it is known that various distributionally robust risk minimization problems with $\phi$-divergence-based or Wasserstein distance-based ambiguity sets can be reformulated as regularized ERM problems~\cite{duchi2019variance,shafieezadeh2019regularization}, the reformulation of the distributionally robust MUGL model~\eqref{eq:min-max} as the regularized ERM problem~\eqref{eq:reform} does not follow from existing results and is new. Moreover, our development above suggests that the (non-squared) Frobenius norm regularizer $\bm{L} \mapsto h_2(\bm{L}) = \rho_2\|\bm{L}\|_F$ is more interpretable than the commonly used squared Frobenius norm regularizer $\bm{L} \mapsto \alpha\|\bm{L}\|_F^2$. Indeed, even though it is often argued that the latter is used to control the sparsity in the learned graph (see, e.g.,~\cite{dong2016learning,mateos2019connecting,berger2020efficient}), such an argument has not been rigorously justified. In addition, as pointed out in~\cite{kalofolias2016learn}, it is not easy to interpret the squared Frobenius norm regularizer, as the elements of $\bm{L}$ are not only of different scales but also linearly dependent.

\section{Solving the MUGL Model} \label{sec-pgd}
Now, let us turn to the algorithmic aspects of the MUGL model~\eqref{eq:reform}. Observe that since $\bm{L} \mapsto \| \bm{L}^{1/2} \widehat{\bm\mu}_n \|_2 = \sqrt{ \widehat{\bm\mu}_n^\top \bm{L} \widehat{\bm\mu}_n }$ is non-convex and even non-Lipschitz at any $\bm{L}$ satisfying $\bm{L}^{1/2} \widehat{\bm\mu}_n = \bm{0}$, the MUGL model~\eqref{eq:reform} gives rise to a challenging non-smooth non-convex optimization problem. Our goal in this section is to develop a PGD method that can efficiently tackle Problem~\eqref{eq:reform} and establish its convergence guarantee.

\subsection{Vectorized MUGL Formulation}\label{subsec-pgd}
Since every Laplacian $\bm{L} \in \mathcal{L}_s$ is symmetric and satisfies $\bm{L}\bm{1}=\bm{0}$, it can be completely specified by, say, the entries below the main diagonal. This motivates us to vectorize Problem~\eqref{eq:reform} to get a more compact formulation. Specifically, let $\bar{m} \coloneqq \tfrac{m(m-1)}{2}$ be the number of entries below the main diagonal of an $m\times m$ matrix and define the linear operator $\mathcal{F}:\mathbb{R}^{\bar{m}} \rightarrow \mathbb{S}^{m}$ by
\begin{equation*}
\begin{split}
\left[\mathcal{F}( \bm{w} )\right]_{ij} \coloneqq
\begin{cases}
-w_{i-j+\frac{j-1}{2}(2m-j)}, &\text{if\ } i>j,\\
\left[\mathcal{F}(\bm{w})\right]_{ji}, &\text{if\ } i<j,\\
-\sum_{k\neq i} \left[\mathcal{F}(\bm{w})\right]_{ik}, &\text{if\ } i=j.
\end{cases}
\end{split}
\end{equation*}
More explicitly, given a vector $\bm{w} \in \R^{\bar{m}}$, the entries below the main diagonal of the matrix $\mathcal{F}(\bm{w})$ are given by
\[ \begin{bmatrix}
 * & * & * & \cdots & * \\
-w_1 & * & * & \cdots & * \\
-w_2 & -w_m & * & \cdots & * \\
\vdots & \vdots & \vdots & \ddots & \vdots \\
-w_{m-1} & -w_{2m-3} & -w_{3m-6} & \cdots & *
\end{bmatrix}
.
\]
Furthermore, define
\[ \Delta_s \coloneqq \left\{ \bm{w} \in \mathbb{R}^{\bar{m}}: \bm{1}^\top \bm{w} = s,\ \bm{w} \geq \bm{0} \right\}. \]
It is not hard to show that $\bm{w} \in \Delta_s$ if and only if $\mathcal{F}(\bm{w}) \in \mathcal{L}_s$. Now, let $\mathcal{F}^*:\mathbb{S}^m \rightarrow \R^{\bar{m}}$ be the adjoint operator of $\mathcal{F}$; i.e., $\mathcal{F}^*$ satisfies $\tr( \mathcal{F}(\bm{w}) \bm{M}) = \bm{w}^\top\mathcal{F}^*(\bm{M})$ for all $\bm{w}\in\R^{\bar{m}}$ and $\bm{M}\in\mathbb{S}^m$. It can be verified that for $1\le j < i \le m$,
\begin{equation}\label{eq-ajoint-def}
\left[\mathcal{F}^*(\bm{M})\right]_{i-j+\frac{j-1}{2}(2m-j)} = M_{ii} - M_{ij} - M_{ji} + M_{jj}.
\end{equation}
Then, Problem~\eqref{eq:reform} admits the vectorized formulation
\begin{equation}\label{eq:reform-vectorize}
		\inf_{\bm{w}\in\Delta_s}  g(\bm{w}) \coloneqq \phi_1(\bm{w}) + \phi_2(\bm{w}) + h(\mathcal{F}(\bm{w})),
\end{equation}
where
\begin{align*}
	\phi_1(\bm{w}) \coloneqq& \ \bm{w}^\top \mathcal{F}^*\left(\widehat{\bm\Sigma}_n+\widehat{\bm\mu}_n\widehat{\bm\mu}_n^\top \right) + \rho_2 \| \mathcal{F}(\bm{w}) \|_F, \\
	\phi_2(\bm{w}) \coloneqq & \ \sqrt{\bm{a}^\top\bm{w}}, \quad \bm{a} \coloneqq 4\rho_1^2 \mathcal{F}^*\left(\widehat{\bm\mu}_n\widehat{\bm\mu}_n^\top\right).
\end{align*}

To have a better understanding of the structure of Problem~\eqref{eq:reform-vectorize}, we first observe that $\bm{w} \mapsto \|\mathcal{F}(\bm{w})\|_F$ is smooth over $\Delta_s$. This follows since $\bm{w} \in \Delta_s$ implies that $\mathcal{F}(\bm{w}) \in \mathcal{L}_s$, which in turn implies that $\|\mathcal{F}(\bm{w})\|_F > 0$. Next, consider the function $\bm{w} \mapsto \phi_2(\bm{w}) = \sqrt{\bm{a}^\top \bm{w}}$. Using~\eqref{eq-ajoint-def}, we deduce that for $1\le j < i \le m$, 
\begin{equation}\label{eq-ak}
\begin{split}
	& a_{i-j+\frac{j-1}{2}(2m-j)} \\
	&= 4\rho_1^2\left( (\widehat{\bm\mu}_n)_i^2 - (\widehat{\bm\mu}_n)_i(\widehat{\bm\mu}_n)_j - (\widehat{\bm\mu}_n)_j(\widehat{\bm\mu}_n)_i + (\widehat{\bm\mu}_n)_j^2\right)\\
	&= 4\rho_1^2\left((\widehat{\bm\mu}_n)_i-(\widehat{\bm\mu}_n)_j\right)^2 \geq 0;
\end{split}
\end{equation}
i.e., $\bm{a} \ge \bm{0}$. Thus, the function $\phi_2$ is well defined on $\Delta_s$ and is non-smooth at any $\bm{w} \in \mathcal{Z}_s \coloneqq \Delta_s \cap \{ \bm{w} \in \R^{\bar{m}} : \bm{a}^\top\bm{w} = 0 \}$. Now, observe that $\mathcal{Z}_s \not= \emptyset$ if and only if there exists a $k \in \{1,\ldots,\bar{m}\}$ such that $a_k=0$, as $\bm{w} \in \Delta_s$ implies that $\bm{w}\ge\bm{0}$ and $\bm{w}\not=\bm{0}$. By~\eqref{eq-ak}, such an event occurs when at least two coordinates of $\widehat{\bm\mu}_n$ are equal. Intuitively, however, if the ground-truth distribution $\bbP^*$ of the graph signal $\bm{x}\in\R^m$ is continuous and $\{\bm{x}^j\}_{j=1}^n$ are $n$ independent realizations of $\bm{x}$, then the probability that $\widehat{\bm\mu}_n$ has at least two equal coordinates should be zero. In other words, the function $\phi_2$ should be smooth on $\Delta_s$ almost surely (i.e., with probability 1). It turns out that such an intuition is almost correct and can be made precise as follows:

\begin{theorem}[Smoothness of MUGL objective] \label{thm-sqrt-smooth}
Let $\bbP^*$ be the ground-truth distribution of the graph signal $\bm{x}\in\R^m$ and $\{\bm{x}^j\}_{j=1}^n$ be $n$ independent realizations of $\bm{x}$. Suppose that $\bbP^*$ is absolutely continuous wrt the $m$-dimensional Lebesgue measure $\nu$ (denoted by $\bbP^* \ll \nu$); i.e., for any measurable set $\mathcal{A} \subseteq \R^m$, $\bbP^*(\mathcal{A})=0$ whenever $\nu(\mathcal{A})=0$. Then, the event $\mathcal{Z}_s = \emptyset$ will occur almost surely. Consequently, the function $\phi_2$ will be smooth on $\Delta_s$ almost surely.
\end{theorem}
\begin{proof}
For $k=1,\ldots,\bar{m}$, define $\mathcal{V}_k \coloneqq \{\bm{u} \in \R^{\bar{m}} : u_k = 0\}$. Recall that $\mathcal{Z}_s \not= \emptyset$ if and only if there exists a $k\in\{1,\ldots,\bar{m}\}$ such that $a_k=0$. Moreover, note that $\bm{a}$ depends on $\bm{x}^1,\ldots,\bm{x}^n$, which are identical and independently distributed according to $\bbP^*$. Thus, we have
	\begin{equation}\label{eq-prob}
		\Pr( \mathcal{Z}_s \not= \emptyset ) = \Pr \left( \bm{a} \in \bigcup_{k=1}^{\bar{m}} \mathcal{V}_k \right) \leq \sum_{k=1}^{\bar{m}}\Pr( \bm{a} \in \mathcal{V}_k ),
	\end{equation}
where the probability is evaluated wrt the product measure ${\bbP^*}^n$. 

Now, consider a fixed $k\in\{1,\ldots,\bar{m}\}$. Then, we can find $i,j \in \{1,\ldots,m\}$ with $1 \le j < i \le m$ such that $k=i-j+\frac{j-1}{2}(2m-j)$. Using~\eqref{eq-ak}, we deduce that
	\begin{equation}\label{eq-a=mu}
			\Pr( \bm{a} \in \mathcal{V}_k ) = \Pr( n\widehat{\bm\mu}_n \in \Pi_{ij} ),
	\end{equation}
where $\Pi_{ij} \coloneqq \{ \bm{v} \in \R^m: v_i = v_j \}$ is an $(m-1)$-dimensional linear subspace in $\R^m$. Since $n\widehat{\bm\mu}_n = \bm{x}^1+\cdots+\bm{x}^n$ is the sum of $n$ independent random vectors that are identically distributed according to $\bbP^*$, its distribution is given by the $n$-fold convolution of $\bbP^*$, denoted by ${\bbP^*}^{\circledast n}$ (cf.~\cite[Section 20]{billingsley2008probability}). Moreover, since $\bbP^* \ll \nu$, we have ${\bbP^*}^{\circledast n} \ll \nu$ (cf.~\cite[Exercise 31.14(b)]{billingsley2008probability}). This, together with the well-known fact that $\nu(\Pi_{ij}) = 0$, implies that 
\begin{equation} \label{eq:0-measure}
\Pr( n\widehat{\bm\mu}_n \in \Pi_{ij} ) = {\bbP^*}^{\circledast n}( \Pi_{ij} ) = 0.
\end{equation}
Upon noting that the above argument holds for arbitrary $k\in\{1,\ldots,\bar{m}\}$ and combining~\eqref{eq-prob}--\eqref{eq:0-measure}, we conclude that
\[ \Pr( \mathcal{Z}_s \not= \emptyset ) \leq \sum_{k=1}^{\bar{m}} \Pr( \bm{a} \in \mathcal{V}_k ) = \sum_{1\le j<i\le m} {\bbP^*}^{\circledast n}( \Pi_{ij} ) = 0. \]
This completes the proof.
\end{proof}

By the Radon--Nikodym theorem~\cite[Section 32]{billingsley2008probability}, the probability distributions that are absolutely continuous wrt the Lebesgue measure are precisely those that have probability density functions wrt the Lebesgue measure. Thus, Theorem~\ref{thm-sqrt-smooth} applies to a wide range of ground-truth distributions. In what follows, we assume that $\bbP^* \ll \nu$.

\subsection{PGD Method for MUGL and Its Convergence Analysis}

Theorem \ref{thm-sqrt-smooth} implies that the gradient of $\phi_2$ at any $\bm{w} \in \Delta_s$ will be well defined almost surely. Thus, for any given $\bm{w} \in \Delta_s$, as long as the gradient of the regularizer $h$ at $\mathcal{F}(\bm{w}) \in \mathcal{L}_s$ is well defined, we can compute the gradient of the objective function $g$ of Problem~\eqref{eq:reform-vectorize} at $\bm{w}$ as follows:
\[ 
	\begin{split}
		\nabla g(\bm{w}) =& \ \mathcal{F}^*\left(\widehat{\bm\Sigma}_n+\widehat{\bm\mu}_n\widehat{\bm\mu}_n^\top\right) + \rho_1 \frac{\mathcal{F}^*\left(\widehat{\bm\mu}_n\widehat{\bm\mu}_n^\top\right)}{\sqrt{\bm{w}^\top \mathcal{F}^*\left(\widehat{\bm\mu}_n\widehat{\bm\mu}_n^\top\right)}} \\
		& \, + \rho_2\frac{\mathcal{F}^*\left(\mathcal{F}(\bm{w})\right)}{\|\mathcal{F}(\bm{w})\|_F} + \mathcal{F}^*(\nabla h(\mathcal{F}(\bm{w}))).
	\end{split}
\] 
This suggests that Problem~\eqref{eq:reform-vectorize} can be tackled by the PGD method, whose update formula is given by
\begin{equation} \label{eq:pgd}
\bm{w}^{k+1} \longleftarrow \Pi_{\Delta_s} ( \bm{w}^k  - \eta_k\nabla g(\bm{w}^k) ), \quad k=0,1,\ldots.
\end{equation}
Here, $\eta_k>0$ is the step size and $\Pi_{\Delta_s}(\bm{w}) \coloneqq \argmin_{\bm{v} \in \Delta_s} \|\bm{v}-\bm{w}\|_2$ is the projection of $\bm{w}$ onto $\Delta_s$. It is well known that the projection of a vector $\bm{w} \in \R^{\bar{m}}$ onto the simplex $\Delta_s$ can be computed in $\mathcal{O}(\bar{m} + {\rm nnz}(\Pi_{\Delta_s}(\bm{w})) \cdot\log\bar{m})$ time, where ${\rm nnz}(\bm{u})$ denotes the number of non-zero elements in $\bm{u}$; see, e.g.,~\cite{condat2016fast} and the references therein. Moreover, the gradient $\nabla g(\bm{w})$ can be computed in $\mathcal{O}(\bar{m})$ time. Thus, the update~\eqref{eq:pgd} can be implemented efficiently.

Since Problem~\eqref{eq:reform-vectorize} is non-convex in general, one does not expect that the PGD method~\eqref{eq:pgd} will find an optimal solution to the problem. Nevertheless, under some mild assumptions, it is possible to establish the convergence of the PGD method~\eqref{eq:pgd} to a stationary point of Problem~\eqref{eq:reform-vectorize}. Recall that a point $\bar{\bm{w}} \in \R^{\bar{m}}$ at which the function $g$ is continuously differentiable is a stationary point of Problem~\eqref{eq:reform-vectorize} if there exists a vector of dual multipliers $(\bar{d}_0,\bar{\bm d}) \in \R^{\bar{m}+1}$ such that $(\bar{\bm{w}};(\bar{d}_0,\bar{\bm d}))$ satisfies the following Karush--Kuhn--Tucker (KKT) conditions:
\begin{align*}
& \nabla g(\bar{\bm{w}}) + \bar{d}_0\bm{1} \ge \bm{0}, \\
& \bm{1}^\top\bar{\bm{w}}=s, \,\,\, \bar{\bm{w}} \ge \bm{0}, \\
& \bar{\bm{w}}^\top(\nabla g(\bar{\bm{w}}) + \bar{d}_0\bm{1}) = 0.
\end{align*}
For notational simplicity, define $\widetilde{g} \coloneqq g + \mathbb{I}_{\Delta_s}$, where $\mathbb{I}_{\Delta_s}$ is the indicator function associated with $\Delta_s$; i.e., $\mathbb{I}_{\Delta_s}(\bm{w})=0$ if $\bm{w}\in\Delta_s$ and $\mathbb{I}_{\Delta_s}(\bm{w})=+\infty$ otherwise. We can now state and prove our first convergence result.
\begin{theorem}[Global convergence of PGD under Lipschitz continuous gradient] \label{thm-conv}
Suppose that the regularizer $h:\mathbb{S}^m \rightarrow \R$ is continuously differentiable and its gradient is Lipschitz continuous on $\mathcal{L}_s$. Then, the following will hold almost surely:
\begin{enumerate}
\item[(a)] The function $g$ is continuously differentiable and its gradient is Lipschitz continuous for some parameter $\ell_g > 0$ on $\Delta_s$.

\item[(b)] If the function $\widetilde{g}$ possesses the \emph{Kurdyka-Łojasiewicz} (KŁ) property (see~\cite[Section 2]{attouch2013convergence} for the definition and a brief discussion of its significance) and the step sizes $\{\eta_k\}_{k\ge0}$ satisfy $\eta_k \in (0,1/\ell_g)$ for all $k\ge0$, then for any initial point $\bm{w}^0 \in \Delta_s$, the sequence $\{\bm{w}^k\}_{k\ge0}$ generated by the PGD method~\eqref{eq:pgd} converges to a stationary point of Problem~\eqref{eq:reform-vectorize}.
\end{enumerate}
\end{theorem}
\begin{proof}
Since $\|\mathcal{F}(\bm{w})\|_F>0$ for all $\bm{w} \in \Delta_s$ and $\Delta_s$ is compact, it is straightforward to show that $\phi_1$ is continuously differentiable and its gradient is Lipschitz continuous on $\Delta_s$. On the other hand, we have $\bbP^* \ll \nu$ by assumption, so that Theorem~\ref{thm-sqrt-smooth} applies. In particular, we will have $\phi_2(\bm{w}) > 0$ for all $\bm{w} \in \Delta_s$ almost surely, which, together with the compactness of $\Delta_s$, implies that $\phi_2$ will be continuously differentiable and its gradient will be Lipschitz continuous on $\Delta_s$ almost surely. These results and the assumption on $h$ immediately yield the result in (a). The result in (b) then follows from a direct application of~\cite[Theorem 5.3]{attouch2013convergence}.
\end{proof}

As discussed in~\cite{attouch2010proximal,attouch2013convergence}, the assumption that $\widetilde{g}$ possesses the KŁ property is a rather mild one. In particular, since $\Delta_s$ is polyhedral and both $\phi_1,\phi_2$ are actually analytic on $\Delta_s$,\footnote{A real-valued function $f:\R^p \rightarrow \R$ is said to be \emph{analytic} on a set $S \subseteq \R^p$ if it is infinitely differentiable at and agrees with its Taylor series in a neighborhood of every point in $S$.} if $h$ is also analytic on $\Delta_s$ (which implies that it satisfies the assumption on $h$ in Theorem~\ref{thm-conv}), then $\widetilde{g}$ possesses the KŁ property; see, e.g., the discussion in~\cite[Section 4.3]{attouch2010proximal}. An important consequence of the KŁ property of $\widetilde{g}$ is that it ensures the \emph{convergence} and not just \emph{subsequential convergence} of the sequence $\{\bm{w}^k\}_{k\ge0}$ generated by the PGD method~\eqref{eq:pgd}.

It is worth noting that $h$ does not have to be convex in order for Theorem~\ref{thm-conv} to hold. On the other hand, if $h$ is convex and $\rho_1=0$ (i.e., there is essentially no uncertainty about the mean of the ground-truth distribution), then~\eqref{eq:reform-vectorize} is a convex optimization problem. In this case, the KKT conditions associated with Problem~\eqref{eq:reform-vectorize} are necessary and sufficient for optimality. Thus, if in addition the assumptions on $h$ and $\widetilde{g}$ in Theorem~\ref{thm-conv} hold, then the iterates generated by the PGD method~\eqref{eq:pgd} converge to an optimal solution to Problem~\eqref{eq:reform-vectorize}.

The convergence result in Theorem~\ref{thm-conv} relies crucially on the Lipschitz continuity of $\nabla g$ on $\Delta_s$. However, for certain choices of the regularizer $h$, the resulting function $g$ may not have such a property. A case in point is the logarithmic barrier regularizer $\mathcal{L}_s \ni \bm{L} \mapsto h_{\rm log}(\bm{L}) \coloneqq -\alpha\sum_{i=1}^m \ln(L_{ii}) \in \R$ with parameter $\alpha>0$, which is introduced in~\cite{kalofolias2016learn} to improve the overall connectivity of the learned graph. Indeed, for any sequence $\{\bm{w}^k\}_{k\ge0}$ in $\Delta_s$ such that $[\mathcal{F}(\bm{w}^k)]_{ii} \rightarrow 0$ for some $i \in \{1,\ldots,m\}$, we have $\| \nabla h_{\rm log}(\mathcal{F}(\bm{w}^k)) \|_2 \rightarrow +\infty$. This implies that the function $g$ cannot have a Lipschitz continuous gradient on $\Delta_s$.

As it turns out, it is possible to ensure the convergence of the PGD method~\eqref{eq:pgd} under weaker smoothness assumptions on $g$ if the step sizes $\{\eta_k\}_{k\ge0}$ are chosen via an appropriate line search strategy. Specifically, let $0<\eta_{\min}\le\eta_{\max}<+\infty$ and $\beta,\gamma \in (0,1)$ be given parameters. Given a sequence $\{\eta_k\}_{k\ge0}$ satisfying $\eta_k \in [\eta_{\min},\eta_{\max}]$, consider a line search-based PGD method (LS-PGD) with the following update scheme:

\smallskip
\begin{subequations} \label{eq:ls-pgd}
For $k=0,1,\ldots$, do the following:
\begin{enumerate}
\item (Projected gradient step). Compute
\begin{align}
\widetilde{\bm w}^{k} &\longleftarrow \Pi_{\Delta_s} ( \bm{w}^k  - \eta_k\nabla g(\bm{w}^k) ), \\
\bm{v}^k &\longleftarrow \widetilde{\bm w}^k - \bm{w}^k.
\end{align}

\item (Armijo-type line search). Compute the least non-negative integer $t$ such that 
\begin{equation} \label{eq:armijo-ls}
\widetilde{g}(\bm{w}^k + \gamma^t \bm{v}^k) \le \widetilde{g}(\bm{w}^k) + \beta\gamma^t \Gamma_k,
\end{equation}
where $\Gamma_k \coloneqq \nabla g(\bm{w}^k)^\top \bm{v}^k + \tfrac{1}{2\eta_k}\|\bm{v}^k\|_2^2$.

\item (Update). Set
\begin{equation} \label{eq:ls-pgd-update}
\bm{w}^{k+1} \longleftarrow \bm{w}^k + \gamma^t\bm{v}^k.
\end{equation}
\end{enumerate}
\end{subequations}
The method described above is a particular instantiation of the one studied in~\cite{bonettini2018block}. However, the convergence guarantees established in~\cite{bonettini2018block} for the method assume that the function $g$ is continuously differentiable on an open set $\Omega_g \subseteq \R^{\bar{m}}$ containing $\Delta_s$. As such, they cannot be directly applied to the setting where the logarithmic barrier regularizer $h_{\rm log}$ is used. Nevertheless, a closer inspection of~\cite{bonettini2018block} reveals that the convergence results therein are still valid if the open set $\Omega_g$ merely satisfies $\Omega_g \cap \Delta_s \not= \emptyset$ and the initial point $\bm{w}^0$ is chosen to lie in $\Omega_g \cap \Delta_s$. This leads to our second convergence result.
\begin{theorem}[Global convergence of LS-PGD under locally Lipschitz continuous gradient] \label{thm:loc-lip-conv}
Let $h:\mathbb{S}^m \rightarrow \R \cup \{+\infty\}$ be a regularizer whose domain ${\rm dom}(h) \coloneqq \{\bm{L} \in \mathbb{S}^m : h(\bm{L}) < +\infty\}$ is open and satisfies ${\rm dom}(h) \cap \mathcal{L}_s \not= \emptyset$. In addition, suppose that $h$ is continuously differentiable on ${\rm dom}(h)$ and its gradient $\nabla h$ is \emph{locally Lipschitz continuous} on ${\rm dom}(h)$ (i.e., for every compact set $\mathcal{B} \subseteq {\rm dom}(h)$, there exists a constant $\ell_{\mathcal{B}}>0$ such that $\| \nabla h(\bm{L}) - \nabla h(\bm{L}') \|_F \le \ell_{\mathcal{B}} \| \bm{L} - \bm{L}' \|_F$ for all $\bm{L},\bm{L}' \in \mathcal{B}$). Then, the following will hold almost surely:
\begin{enumerate}
\item[(a)] The function $g$ is continuously differentiable on an open set $\Omega_g \subseteq \R^{\bar{m}}$ with $\Omega_g \cap \Delta_s \not= \emptyset$ and its gradient $\nabla g$ is locally Lipschitz continuous on $\Omega_g$.

\item[(b)] If the function $\widetilde{g}$ possesses the KŁ property, then for any initial point $\bm{w}^0 \in \Omega_g \cap \Delta_s$, the sequence $\{\bm{w}^k\}_{k\ge0}$ generated by the LS-PGD method~\eqref{eq:ls-pgd} converges to a stationary point of Problem~\eqref{eq:reform-vectorize}.
\end{enumerate}
\end{theorem}
\begin{proof}
For any $\mathcal{U} \subseteq \mathbb{S}^m$, define $\mathcal{F}^{-1}(\mathcal{U}) \coloneqq \{ \bm{w} \in \R^{\bar{m}} : \mathcal{F}(\bm{w}) \in \mathcal{U} \}$.
Observe that
\begin{align*}
\mathcal{F}^{-1}({\rm dom}(h) \cap \mathcal{L}_s) &= \mathcal{F}^{-1}({\rm dom}(h)) \cap \mathcal{F}^{-1}(\mathcal{L}_s) \\
&= \mathcal{F}^{-1}({\rm dom}(h)) \cap \Delta_s.
\end{align*}
This, together with the assumption on $h$ and the continuity of $\mathcal{F}$, implies that $\bm{w} \mapsto h(\mathcal{F}(\bm{w}))$ is continuously differentiable on the open set $\mathcal{F}^{-1}({\rm dom}(h))$ with $\mathcal{F}^{-1}({\rm dom}(h)) \cap \Delta_s \not= \emptyset$. On the other hand, note that the function $\phi_1$ is continuously differentiable on a bounded open set $\Xi_1$ that contains $\Delta_s$. Moreover, since $\bbP^* \ll \nu$ by assumption, Theorem~\ref{thm-sqrt-smooth} implies that almost surely, the function $\phi_2$ will be continuously differentiable on a bounded open set $\Xi_2$ that contains $\Delta_s$. By taking $\Omega_g = \mathcal{F}^{-1}({\rm dom}(h)) \cap \Xi_1 \cap \Xi_2$, we see that $\Omega_g$ is open and $g$ is continuously differentiable on $\Omega_g$ with $\Omega_g \cap \Delta_s = \mathcal{F}^{-1}({\rm dom}(h)) \cap \Delta_s \not= \emptyset$. Lastly, the assumption that $\nabla h$ is locally Lipschitz continuous on ${\rm dom}(h)$, together with the fact that $\nabla\phi_1$ and $\nabla\phi_2$ are Lipschitz continuous on $\Xi_1$ and $\Xi_2$, respectively due to the boundedness of $\Xi_1$ and $\Xi_2$, implies that $\nabla g$ is locally Lipschitz continuous on $\Omega_g$. This establishes the result in (a).

Now, let us prove by induction that $\bm{w}^k \in \Omega_g \cap \Delta_s$ for all $k\ge0$. The base case follows from our assumption. For the inductive step, we first note that by the convexity of $\Delta_s$, we have $\bm{w}^k + \gamma^t\bm{v}^k = (1-\gamma^t)\bm{w}^k + \gamma^t \widetilde{\bm w}^k \in \Delta_s$ for all $t\ge0$. This, together with the fact that $\Delta_s \subset \Xi_i$ for $i=1,2$, implies that if $\bm{w}^k + \gamma^{t'}\bm{v}^k \not \in \Omega_g$ for some $t'\ge0$, then $\widetilde{g}(\bm{w}^k + \gamma^{t'}\bm{v}^k) = +\infty$; i.e., condition~\eqref{eq:armijo-ls} is not satisfied. Next, note that since $\Omega_g$ is open with $\bm{w}^k \in \Omega_g$ and $\gamma \in (0,1)$, there exists an integer $T\ge0$ such that $\bm{w}^k + \gamma^t\bm{v}^k \in \Omega_g$ for all $t\ge T$. Since the line search in the LS-PGD method~\eqref{eq:ls-pgd} terminates in a finite number of steps (see the discussion in~\cite[Section 3.1]{bonettini2018block}), we conclude that $\bm{w}^{k+1} \in \Omega_g \cap \Delta_s$, which completes the inductive step. The result in (b) can then be obtained by following the development in~\cite[Section 3.3]{bonettini2018block}.
\end{proof}

The assumption on the regularizer $h$ in Theorem~\ref{thm:loc-lip-conv} is much milder than that in Theorem~\ref{thm-conv}. In particular, Theorem~\ref{thm:loc-lip-conv} applies to the setting where the logarithmic barrier regularizer $h_{\rm log}$ is used. Under such a setting, the KŁ property of $\widetilde{g}$ follows from the analyticity of $\bm{w} \mapsto h_{\rm log}(\mathcal{F}(\bm{w}))$ on the open set $\mathcal{F}^{-1}({\rm dom}(h_{\rm log})) = \{ \bm{w} \in \R^{\bar{m}} :  [\mathcal{F}(\bm{w})]_{ii} > 0 \mbox{ for } i=1,\ldots,m \}$, the analyticity of $\phi_1,\phi_2$ on a bounded open set containing $\Delta_s$, and the polyhedrality of $\Delta_s$. As an aside, let us point out that the LS-PGD method~\eqref{eq:ls-pgd} can also be used to solve the non-robust graph learning model proposed in~\cite{kalofolias2016learn}, which is an instance of Problem~\eqref{eq:gl-regular} with $h(\bm{L}) = - \alpha \sum_{i=1}^m \ln(L_{ii}) + \tfrac{\beta}{2} \sum_{1\le i\not=j \le m} L_{ij}^2$ and can be equivalently written as
\begin{equation} \label{eq:non-robust-log-barrier}
\inf_{\bm{w} \in \Delta_s} \left\{ \phi(\bm{w}) + h_{\rm log}(\mathcal{F}(\bm{w})) \right\}
\end{equation}
with 
\[ \phi(\bm{w}) \coloneqq \bm{w}^\top \mathcal{F}^*\left(\widehat{\bm\Sigma}_n + \widehat{\bm\mu}_n\widehat{\bm\mu}_n^\top \right) + \frac{\beta}{2} \sum_{1\le i\not=j \le m} [\mathcal{F}(\bm{w})]_{ij}^2. \]
Since Problem~\eqref{eq:non-robust-log-barrier} has a strongly convex objective function and a convex feasible set, Theorem~\ref{thm:loc-lip-conv} guarantees that the iterates generated by the method converge globally to its unique optimal solution. Interestingly, in the context of solving the non-robust graph learning model~\eqref{eq:non-robust-log-barrier}, both the proposed LS-PGD method~\eqref{eq:ls-pgd} and its convergence guarantee given in Theorem~\ref{thm:loc-lip-conv} are new and can be of independent interest; cf.~\cite{wang2021efficient}.


\section{Numerical Results}\label{sec-exp}
In this section, we study the performance of our proposed distributionally robust MUGL model and several representative non-robust graph learning models in the literature via numerical experiments on both synthetic and real-world data. Specifically, for the distributionally robust MUGL model, we consider the formulation~\eqref{eq:reform} with $s=m$, $h(\bm{L})=0$ (denoted by MUGL-\textit{o}) and $s=m$, $h(\bm{L})=-\alpha\sum_{i=1}^m \ln(L_{ii})$ (denoted by MUGL-\textit{l}). For the non-robust graph learning models, we consider the following formulations:
\begin{itemize}
	\item Vanilla smooth graph learning (VSGL) model:
	\[ 
		\inf_{\bm{L}\in\mathcal{L}_m} \tr(\bm{X}^\top \bm{L} \bm{X}).
	\] 
	\item GL-SigRep model \cite{dong2016learning} with parameters $\beta,\gamma>0$:
	\[ 
			\inf_{\substack{ \bm{L} \in \mathcal{L}_{m} \\ \bm{Y} \in \R^{m\times n}}} \left\{ \|\bm{X}-\bm{Y}\|_F^2 + \beta\tr( \bm{Y}^\top \bm{L} \bm{Y} ) +\gamma \|\bm{L}\|_F^2 \right\}.
	\] 
	\item Log-barrier model \cite{kalofolias2016learn,kalofolias2018large} with parameters $\beta,\gamma>0$:
	\[ 
		\begin{split}
			\inf_{\bm{W} \in \mathbb{S}^m} & \left\{ \frac{1}{2} \tr(\bm{Z} \bm{W}) - \beta\sum_{i=1}^m \ln\left( \sum_{j=1}^m W_{ij} \right) + \frac{\gamma}{2} \|\bm{W}\|_F^2 \right\} \\
			\text{s.t.}\,\,\,\, &\ \bm{W} \geq \bm{0}, \, W_{ii} = 0 \mbox{ for } i=1,\ldots,m,
		\end{split}
	\] 
	where $\bm{Z} \in \mathbb{S}^m$ is the pairwise distance matrix given by $Z_{jk} = \| \bm{x}^j - \bm{x}^k \|_2^2$ for $j,k = 1,\ldots,m$.
\end{itemize}
%
To evaluate the performance of the different models, we use the following metrics, which are standard in the data science literature (see, e.g., \cite{manning2008introduction}): 
\begin{align*}
	&\text{F-measure}=\frac{2{\sf TP}}{2{\sf TP}+{\sf FN}+{\sf FP}},\\
	&\text{Precision}=\frac{\sf TP}{{\sf TP}+{\sf FP}}, \quad \text{Recall}=\frac{\sf TP}{{\sf TP}+{\sf FN}}, \\
	&\text{Normalized Mutual Information (NMI)} \\
	&\,\,\,=\frac{2\times I({\sf TP}+{\sf FN}; {\sf TP}+{\sf FP})}{H({\sf TP}+{\sf FN})+H({\sf TP}+{\sf FP})}.
\end{align*}
Here, ${\sf TP}$, ${\sf FP}$, and ${\sf FN}$ denote the number of true positives, false positives, and false negatives, respectively; $H({\sf TP}+{\sf FN})$ and $H({\sf TP}+{\sf FP})$ denote the entropy of the edges in the underlying graph and in the learned graph, respectively; $I({\sf TP}+{\sf FN}; {\sf TP}+{\sf FP})$ denotes the mutual information between the edges in the underlying graph and those in the learned graph. A learning algorithm is deemed good if it achieves a high F-measure or NMI value. Note that F-measure is the harmonic mean of precision and recall. Thus, a good learning algorithm should in principle achieve high precision and recall values simultaneously.

Both the MUGL-\textit{o} and VSGL models can be solved using the PGD method~\eqref{eq:pgd}, while the MUGL-\textit{l} model can be solved using the LS-PGD method~\eqref{eq:ls-pgd}. We initialize both the PGD~\eqref{eq:pgd} and LS-PGD~\eqref{eq:ls-pgd} methods by the centroid of $\Delta_m$. We use the code provided by the authors of~\cite{dong2016learning},\footnote{\url{http://web.media.mit.edu/~xdong/code/graphlearning.zip}} which implements an alternating minimization method, to solve the GL-SigRep model. We use the code provided in the Graph Signal Processing Toolbox~\cite{perraudin2014gspbox},\footnote{\url{https://epfl-lts2.github.io/gspbox-html/doc/demos/gsp_demo_learn_graph_large.html}} which implements a primal-dual method with the tricks given in~\cite{kalofolias2018large}, to solve the Log-barrier model. All reported results are obtained using the best-tuned model parameters, so that the learned graphs have the highest quality in terms of F-measure. 

	\begin{table*}
		\begin{subtable}{0.5\linewidth}
			\fontsize{7.5}{11}\selectfont
			\centering
			\setlength\tabcolsep{1.5pt}
			\begin{tabular}{l|l|l|l|l|l}
				\hline
				\multicolumn{2}{c|}{} & \multicolumn{1}{c|}{F-measure} & \multicolumn{1}{c|}{Precision}  & \multicolumn{1}{c|}{Recall} & \multicolumn{1}{c}{NMI} \\
				\hline
				\multirow{5}{*}{\rotatebox[origin=c]{90}{Gaussian}} 
				&VSGL & 0.411$\pm$9.87\% & \textbf{0.963$\pm$5.07\%} & 0.262$\pm$12.46\% & 0.212$\pm$17.57\% \\
				&GL-SigRep & 0.740$\pm$4.99\% & 0.593$\pm$8.00\% & \textbf{0.988$\pm$1.30\%} & 0.401$\pm$13.39\% \\
				&Log-Model & 0.756$\pm$5.53\% & 0.850$\pm$5.14\% & 0.683$\pm$8.01\% & 0.396$\pm$16.05\% \\
				&MUGL-\textit{o} & 0.779$\pm$5.21\% & 0.831$\pm$5.38\% & 0.734$\pm$7.05\% & 0.415$\pm$16.26\% \\
				&MUGL-\textit{l} & \textbf{0.785$\pm$5.70\%} & 0.816$\pm$6.46\% & 0.758$\pm$6.91\% & \textbf{0.418$\pm$19.08\%} \\
				\hline
				\multirow{5}{*}{\rotatebox[origin=c]{90}{ER}} 
				&VSGL & 0.413$\pm$14.95\% & \textbf{0.663$\pm$13.77\%} & 0.302$\pm$17.33\% & 0.146$\pm$33.57\% \\
				&GL-SigRep & 0.580$\pm$7.26\% & 0.462$\pm$8.84\% & 0.784$\pm$7.99\% & 0.213$\pm$22.77\% \\
				&Log-Model & \textbf{0.620$\pm$8.28}\% & 0.550$\pm$9.21\% & 0.712$\pm$9.23\% & \textbf{0.248$\pm$23.77\%} \\
				&MUGL-\textit{o} & 0.554$\pm$6.45\% & 0.416$\pm$8.34\% & 0.832$\pm$5.99\% & 0.195$\pm$20.63\% \\
				&MUGL-\textit{l} & 0.570$\pm$6.49\% & 0.424$\pm$8.35\% & \textbf{0.876$\pm$5.34\%} & 0.223$\pm$20.73\% \\
				\hline
				\multirow{5}{*}{\rotatebox[origin=c]{90}{PA}} 
				&VSGL & 0.546$\pm$7.46\% & 0.386$\pm$9.76\% & 0.938$\pm$5.16\% & 0.294$\pm$17.61\% \\
				&GL-SigRep & 0.556$\pm$7.75\% & 0.414$\pm$10.02\% & 0.851$\pm$7.43\% & 0.271$\pm$18.11\% \\
				&Log-Model & \textbf{0.654$\pm$7.86\%} & \textbf{0.679$\pm$10.09\%} & 0.636$\pm$9.87\% & \textbf{0.359$\pm$17.06\%} \\
				&MUGL-\textit{o} & 0.567$\pm$6.68\% & 0.400$\pm$9.28\% & \textbf{0.981$\pm$2.82\%} & 0.337$\pm$13.00\% \\
				&MUGL-\textit{l} & 0.571$\pm$6.50\% & 0.405$\pm$9.15\% &  0.972$\pm$4.13\% & 0.336$\pm$13.63\% \\ 
				\hline
			\end{tabular}
			\vspace*{-1mm}
			\caption{$m=20, n=30, \epsilon=0.1$}
			\vspace*{2mm}
			\label{tab-m20-n30-noise0.1}
		\end{subtable}\hfill
		\begin{subtable}{0.5\linewidth}
			\fontsize{7.5}{11}\selectfont
			\centering
			\setlength\tabcolsep{1.5pt}
			\begin{tabular}{l|l|l|l|l|l}
				\hline
				\multicolumn{2}{c|}{} & \multicolumn{1}{c|}{F-measure} & \multicolumn{1}{c|}{Precision}  & \multicolumn{1}{c|}{Recall} & \multicolumn{1}{c}{NMI} \\
				\hline
				\multirow{5}{*}{\rotatebox[origin=c]{90}{Gaussian}} 
				&VSGL & 0.473$\pm$8.51\% & \textbf{0.987$\pm$2.63\%} & 0.312$\pm$11.24\% & 0.264$\pm$12.55\% \\
				&GL-SigRep & 0.791$\pm$5.78\% & 0.788$\pm$3.65\% & 0.799$\pm$10.66\% & 0.423$\pm$16.21\% \\
				&Log-Model & 0.773$\pm$5.23\% & 0.879$\pm$4.03\% & 0.692$\pm$7.98\% & 0.430$\pm$14.16\% \\
				&MUGL-\textit{o} & 0.830$\pm$3.08\% & 0.856$\pm$2.46\% & 0.807$\pm$4.45\% & 0.497$\pm$9.75\% \\
				&MUGL-\textit{l} & \textbf{0.840$\pm$3.84\%} & 0.842$\pm$3.91\% & \textbf{0.842$\pm$7.33\%} & \textbf{0.517$\pm$12.20\%} \\
				\hline
				\multirow{5}{*}{\rotatebox[origin=c]{90}{ER}}
				&VSGL & 0.493$\pm$11.27\% & 0.763$\pm$10.22\% & 0.366$\pm$13.88\% & 0.216$\pm$25.59\% \\
				&GL-SigRep & 0.512$\pm$9.19\% & \textbf{0.767$\pm$11.55\%} & 0.387$\pm$11.54\% & 0.228$\pm$23.01\% \\
				&Log-Model & 0.560$\pm$8.83\% & 0.591$\pm$9.13\% & 0.535$\pm$10.85\% & 0.201$\pm$23.68\% \\
				&MUGL-\textit{o}  & 0.545$\pm$9.49\% & 0.752$\pm$8.07\% & 0.429$\pm$12.39\% & \textbf{0.241$\pm$20.14\%}  \\
				&MUGL-\textit{l} & \textbf{0.584$\pm$8.90\%} & 0.560$\pm$9.45\% &   \textbf{0.613$\pm$11.01\%} & 0.214$\pm$26.01\% \\
				\hline
				\multirow{5}{*}{\rotatebox[origin=c]{90}{PA}} 
				&VSGL & 0.715$\pm$7.39\% & \textbf{0.994$\pm$2.27\%} & 0.561$\pm$10.99\% & 0.541$\pm$10.98\% \\ 
				&GL-SigRep & 0.557$\pm$5.19\% & 0.561$\pm$11.25\% & 0.562$\pm$10.09\% & 0.249$\pm$12.06\% \\
				&Log-Model & 0.747$\pm$4.72\% & 0.765$\pm$6.27\% & 0.732$\pm$5.86\% &  0.477$\pm$10.57\%\\
				&MUGL-\textit{o} & 0.826$\pm$5.35\% & 0.905$\pm$6.17\% & 0.764$\pm$8.83\% & 0.620$\pm$11.64\% \\ 
				&MUGL-\textit{l} & \textbf{0.893$\pm$5.44\%} & 0.963$\pm$3.71\% & \textbf{0.834$\pm$7.80\%} & \textbf{0.747$\pm$12.50\%} \\ 
				\hline
			\end{tabular}
			\vspace*{-1mm}
			\caption{$m=20, n=80, \epsilon=0.1$}
			\vspace*{2mm}
			\label{tab-m20-n80-noise0.1}
		\end{subtable}
		\begin{subtable}{0.5\linewidth}
			\fontsize{7.5}{11}\selectfont
			\centering
			\setlength\tabcolsep{1.5pt}
			\begin{tabular}{l|l|l|l|l|l}
				\hline
				\multicolumn{2}{c|}{} & \multicolumn{1}{c|}{F-measure} & \multicolumn{1}{c|}{Precision}  & \multicolumn{1}{c|}{Recall} & \multicolumn{1}{c}{NMI} \\
				\hline
				\multirow{5}{*}{\rotatebox[origin=c]{90}{Gaussian}} 
				&VSGL & 0.222$\pm$21.30\% & \textbf{0.867$\pm$14.30\%} & 0.128$\pm$23.33\% & 0.092$\pm$44.54\% \\
				&GL-SigRep & 0.653$\pm$18.30\% & 0.521$\pm$25.83\% & \textbf{0.919$\pm$6.98\%} & 0.241$\pm$59.03\% \\
				&Log-Model & 0.626$\pm$8.37\% & 0.730$\pm$8.76\% & 0.550$\pm$10.32\% & 0.225$\pm$28.22\% \\
				&MUGL-\textit{o} & 0.661$\pm$8.20\% & 0.670$\pm$10.62\% & 0.656$\pm$8.03\% & 0.231$\pm$30.33\%  \\
				&MUGL-\textit{l} & \textbf{0.700$\pm$7.09\%} & 0.608$\pm$9.86\% & 0.831$\pm$7.11\% & \textbf{0.272$\pm$27.34\%} \\
				\hline
				\multirow{5}{*}{\rotatebox[origin=c]{90}{ER}} 
				&VSGL & 0.182$\pm$34.18\% & \textbf{0.461$\pm$29.80\%} & 0.115$\pm$37.22\% & 0.037$\pm$68.13\%  \\
				&GL-SigRep & 0.354$\pm$11.66\% & 0.257$\pm$16.71\% & 0.645$\pm$29.52\% & 0.035$\pm$55.10\% \\
				&Log-Model & 0.490$\pm$8.50\% & 0.342$\pm$9.68\%  & \textbf{0.863$\pm$8.06\%}  & \textbf{0.148$\pm$31.94\%} \\
				&MUGL-\textit{o} & 0.438$\pm$5.50\% & 0.355$\pm$14.98\% & 0.576$\pm$11.51\% & 0.091$\pm$46.28\%  \\
				&MUGL-\textit{l} & \textbf{0.510$\pm$9.67\%} & 0.408$\pm$13.21\% &    0.692$\pm$10.04\% & 0.139$\pm$33.37\% \\
				\hline
				\multirow{5}{*}{\rotatebox[origin=c]{90}{PA}} 
				&VSGL & 0.297$\pm$40.12\% & \textbf{0.558$\pm$37.85\%} & 0.205$\pm$43.42\% & 0.120$\pm$72.23\%  \\
				&GL-SigRep & 0.365$\pm$22.03\% & 0.418$\pm$25.05\% & 0.328$\pm$22.33\% & 0.109$\pm$51.04\% \\
				&Log-Model  &   0.424$\pm$21.77\% & 0.426$\pm$20.12\%      & 0.424$\pm$20.26\% &    0.139$\pm$42.90\% \\
				&MUGL-\textit{o} & 0.436$\pm$15.82\% & 0.331$\pm$17.56\% & \textbf{0.643$\pm$15.09\%} & 0.141$\pm$40.08\%  \\
				&MUGL-\textit{l} & \textbf{0.460$\pm$25.93\%} & 0.490$\pm$19.72\% & 0.434$\pm$21.59\% & \textbf{0.180$\pm$54.72\%} \\ 
				\hline
			\end{tabular}
			\vspace*{-1mm}
			\caption{$m=20, n=30, \epsilon=1$}
			\vspace*{2mm}
			\label{tab-m20-n30-noise1}
		\end{subtable}\hfill
		\begin{subtable}{0.5\linewidth}
			\fontsize{7.5}{11}\selectfont
			\centering
			\setlength\tabcolsep{1.5pt}
			\begin{tabular}{l|l|l|l|l|l}
				\hline
				\multicolumn{2}{c|}{} & \multicolumn{1}{c|}{F-measure} & \multicolumn{1}{c|}{Precision}  & \multicolumn{1}{c|}{Recall} & \multicolumn{1}{c}{NMI} \\
				\hline
				\multirow{5}{*}{\rotatebox[origin=c]{90}{Gaussian}} 
				&VSGL & 0.279$\pm$20.63\% & \textbf{0.929$\pm$8.62\%} & 0.165$\pm$23.85\% & 0.133$\pm$34.70\% \\
				&GL-SigRep & 0.772$\pm$8.21\% & 0.757$\pm$7.46\% & 0.794$\pm$12.25\% & 0.393$\pm$23.90\% \\
				&Log-Model & 0.718$\pm$7.02\% & 0.817$\pm$6.66\% & 0.641$\pm$8.85\% & 0.340$\pm$21.29\% \\
				&MUGL-\textit{o} & 0.756$\pm$4.98\% & 0.637$\pm$8.67\% & \textbf{0.938$\pm$4.38\%} & 0.387$\pm$16.27\% \\
				&MUGL-\textit{l} & \textbf{0.856$\pm$2.46\%} & 0.807$\pm$4.45\% & 0.830$\pm$3.08\% & \textbf{0.497$\pm$9.75\%} \\
				\hline
				\multirow{5}{*}{\rotatebox[origin=c]{90}{ER}} 
				&VSGL & 0.265$\pm$27.30\% & \textbf{0.573$\pm$23.03\%} & 0.174$\pm$30.03\% & 0.074$\pm$56.41\% \\
				&GL-SigRep & 0.373$\pm$6.95\% & 0.252$\pm$12.70\% & \textbf{0.752$\pm$14.49\%} & 0.042$\pm$34.87\% \\
				&Log-Model & 0.446$\pm$18.05\% & 0.554$\pm$18.68\% & 0.375$\pm$19.05\% & 0.133$\pm$46.00\% \\
				&MUGL-\textit{o} & 0.441$\pm$15.8\% & 0.541$\pm$15.32\%  & 0.376$\pm$18.36\% & 0.125$\pm$39.25\% \\
				&MUGL-\textit{l} & \textbf{0.508$\pm$9.18\%} & 0.410$\pm$13.30\% & 0.676$\pm$8.80\% & \textbf{0.135$\pm$31.58\%} \\
				\hline
				\multirow{5}{*}{\rotatebox[origin=c]{90}{PA}} 
				&VSGL & 0.454$\pm$23.57\% & \textbf{0.827$\pm$15.89\%} & 0.320$\pm$29.69\% & 0.261$\pm$37.10\% \\
				&GL-SigRep & 0.494$\pm$15.17\% & 0.762$\pm$16.97\% & 0.369$\pm$17.62\% & 0.263$\pm$31.47\% \\
				&Log-Model & 0.534$\pm$12.43\% & 0.579$\pm$12.88\% & 0.497$\pm$13.05\% & 0.236$\pm$27.16\% \\
				&MUGL-\textit{o} & 0.537$\pm$18.57\% & 0.789$\pm$17.31\% & 0.413$\pm$22.07\% & \textbf{0.304$\pm$34.17\%} \\
				&MUGL-\textit{l} & \textbf{0.557$\pm$11.09\%} & 0.476$\pm$13.41\% & \textbf{0.678$\pm$12.71\%} & 0.247$\pm$26.52\% \\
				\hline
			\end{tabular}
			\vspace*{-1mm}
			\caption{$m=20, n=80, \epsilon=1$}
			\vspace*{2mm}
			\label{tab-m20-n80-noise1}
		\end{subtable}
		\vspace{-2mm}
		\caption{Prediction performance on synthetic data with different values of $m$, $n$, and $\epsilon$}
		\label{tab:synthetic}
	\end{table*}

\vspace{-0.15cm}
\subsection{Experiments on Synthetic Data}
We conduct experiments on three types of synthetic graphs, namely, the Gaussian graph, the Erd{\H{o}}s-R{\'e}nyi (ER) graph, and the preferential attachment (PA) graph. The Gaussian graphs used in our experiments are generated as follows: First, the nodes are placed uniformly at random in a unit square. Then, an edge is placed between nodes $i$ and $j$ ($i \not= j$) if the weight determined by the radial basis function $\exp\left(-d(i,j)^2/2\sigma^2\right)$, where $d(i,j)$ is the Euclidean distance between nodes $i$ and $j$ and $\sigma=0.5$ is the kernel width parameter, is at least $0.75$. The ER graphs are generated by placing an edge between each pair of nodes independently with probability $p=0.2$. The PA graphs are generated by having $\theta_0=2$ connected nodes initially and then adding new nodes one at a time, where each new node is connected to exactly $\theta=1$ previous node that is randomly chosen with a probability proportional to its degree at the time. The edges in the Gaussian graph have weights given by the radial basis function, while those in the ER and PA graphs are set to 1. After obtaining the synthetic graphs, we use the factor analysis model introduced in~\cite{dong2016learning} to generate the graph signals on them. Specifically, given a graph, let $\bm{L}^*$ be its Laplacian whose eigen-decomposition is given by $\bm{L}^*=\bm{\chi}\bm{\Lambda}\bm{\chi}^\top$.
The graph signal $\bm{x} \in \mathbb{R}^{m}$ is then generated according to \eqref{eq-signal-gen}.

To evaluate the efficacy of the different graph learning models, we consider two noise levels $\epsilon = 0.1$ and $1$ in the factor analysis model~\eqref{eq-signal-gen}. We examine the performance of the graphs learned from $n=30$ or $80$ independently generated noisy graph signals, averaged over 50 runs. The results are reported in Table~\ref{tab:synthetic}, where the entries give the average performance as measured by the stated metrics and the associated normalized standard deviations. We observe that VSGL always yields the highest precision values, since this vanilla model usually produces very sparse graphs. Moreover, MUGL-\textit{o} exhibits significant performance gain over VSGL in terms of F-measure and NMI. This demonstrates the advantage of robustifying the vanilla graph learning model against moment uncertainties about the ground-truth distribution. 

As can also be observed from the results, MUGL-\textit{l} achieves the highest F-measure and NMI values in most cases. Moreover, the normalized standard deviation of MUGL is generally lower than those of the other three non-robust methods, particularly in the experiments involving Gaussian graphs with relatively noisy ($\epsilon=1$) graph signals. This indicates that our proposed MUGL model is able to achieve its main aim of attaining a more consistent performance across different populations of observed signals.

\vspace{-0.15cm}
\subsection{Experiments on Real Temperature Data}
We conduct experiments on the real-world temperature data provided in \cite{dong2016learning}. The dataset consists of monthly temperature data from 1981 to 2010 collected by 89 measuring stations in Switzerland. We construct a graph in which the nodes correspond to the measuring stations and the edges correspond to two stations whose altitude difference is less than 300 meters. Furthermore, we assign a weight of 1 to each edge. Such a construction is motivated by the fact that temperature difference is highly related to altitude difference. For each station, we compute the average temperature of each month over the 30-year period. Thus, each month yields a graph signal, and we have 12 graph signals in total. Given these signals, we aim to recover the graph that reflects the altitude relationships between the stations. We then evaluate the learned graph using the same metrics as those in the previous subsection. The results are reported in Table \ref{tab-temperature}. VSGL performs fairly well in this scenario, but MUGL-\textit{o} performs even better in terms of F-measure and NMI. Among the five considered approaches, MUGL-\textit{l} achieves the highest F-measure and NMI values.

\begin{table}
	\fontsize{9}{12}\selectfont
	\centering
	\begin{tabular}{c|cccccc}
		\hline
		& F-measure & Precision & Recall & NMI\\
		\hline
		VSGL & 0.744 & 0.597 & \textbf{0.986} & 0.389\\
		\hline
		GL-SigRep & 0.791 & 0.828 & 0.757 & 0.415\\
		\hline
		Log-Model & 0.759 & \textbf{0.894} & 0.660 & 0.411\\
		\hline
		MUGL-\textit{o} & 0.827 & 0.736 & 0.943 & 0.488 \\
		\hline
		MUGL-\textit{l} & \textbf{0.837} & 0.802 & 0.875 & \textbf{0.504}\\
		\hline
	\end{tabular}
	\caption{Prediction performance on real temperature data}
	\label{tab-temperature}
\end{table}

\vspace{-0.15cm}
\subsection{Experiments on Real Image Data}
We further evaluate the efficacy of the different models by applying them to learn the similarity graph of real images and using the learned graph to perform spectral clustering. We consider two different image datasets, namely, USPS \cite{hull1994database} and COIL-20 \cite{nene1996columbia}. The former consists of 7291 training images and 2007 test images, each of which is a $16 \times 16$ grayscale handwritten digit from 0 to 9. The latter consists of 1440 images of 20 different objects, each of which is downsampled to a size of $32 \times 32$. In the context of clustering, these two image datasets contain 10 and 20 clusters, respectively. 

Spectral clustering \cite{von2007tutorial} aims to perform dimensionality reduction based on the eigenvalues of a so-called \emph{similarity matrix} of the data. The low-dimensional spectral representations of the original data are then clustered using standard methods such as $k$-means. The similarity matrix captures the relative similarity of each pair of points in the dataset in a quantitative manner, and its quality will influence the performance of spectral clustering.
Given an image dataset, we postulate that there is an unknown complete $m$-vertex graph in which (i) each node corresponds to an image in the dataset and (ii) the weight of an edge between two nodes represents the similarity between the two corresponding images. Each pixel of an image gives an observed value at the node corresponding to that image. The collection of all such values constitutes our graph signals. In particular, if the images in a dataset are of size $n_r\times n_c$, then there are $n=n_r\times n_c$ graph signals. Our goal then is to learn a similarity matrix from these graph signals, so as to facilitate the subsequent clustering task.

In each run of the experiment, we randomly pick 100 images from the USPS dataset and 200 images from the COIL-20 dataset, which give rise to graphs with $m=100$ and $m=200$ vertices, respectively. According to the sizes of the images in the USPS and COIL-20 datasets, we obtain $n = 16 \times 16 = 256$ graph signals from the former and $n = 32 \times 32 = 1024$ graph signals from the latter. Given the graphs learned by different models, we run the spectral clustering algorithm in~\cite{ng2001spectral} and evaluate the results using the following common clustering performance metrics: Jaccard coefficient (JC), Fowlkes and Mallows index (FMI), and Rand index (RI) \cite{han2011data}. All these metrics yield values that lie in $[0,1]$, and a higher value indicates better performance in principle. The results are presented in Table \ref{tab-sc}, where the entries give the average performance as measured by the stated metrics and the associated normalized standard deviations over 10 runs. We do not report the results of GL-SigRep, since it is solved based on the CVX solver, which is hardly scalable to medium-sized problems. Among the compared approaches, MUGL-\textit{o} and MUGL-\textit{l} achieve better and more consistent performance in terms of all three aforementioned metrics in both the USPS and COIL-20 datasets. This indicates the high quality and robustness of the learned graphs produced by the MUGL model.

	\begin{table}
		\begin{subtable}[t]{0.48\textwidth}
			\fontsize{9}{12}\selectfont
			\centering
			\setlength\tabcolsep{3pt}
			\begin{tabular}{l|l|l|l}
				\hline
				& \multicolumn{1}{c|}{JC} & \multicolumn{1}{c|}{FMI} & \multicolumn{1}{c}{RI} \\
				\hline
				VSGL & 0.138$\pm$21.78\% & 0.321$\pm$13.99\% & 0.512$\pm$20.56\% \\
				Log-Model & 0.100$\pm$3.99\% & 0.243$\pm$6.30\% & 0.476$\pm$12.48\% \\
				MUGL-\textit{o} & 0.333$\pm$15.18\% & 0.498$\pm$11.33\% & 0.892$\pm$1.67\% \\
				MUGL-\textit{l} & \textbf{0.341$\pm$16.45\%} & \textbf{0.507$\pm$12.25\%} & \textbf{0.896$\pm$1.64\%} \\
				\hline
			\end{tabular}
			\caption{USPS}
			\label{tab-usps-sc}	
		\end{subtable}
		\bigskip
		\vspace{0.0cm}
		\begin{subtable}[t]{0.48\textwidth}
			\fontsize{9}{12}\selectfont
			\centering
			\setlength\tabcolsep{3pt}
			\begin{tabular}{l|l|l|l}
				\hline
				& \multicolumn{1}{c|}{JC} & \multicolumn{1}{c|}{FMI} & \multicolumn{1}{c}{RI} \\
				\hline
				VSGL & 0.355$\pm$12.05 & 0.526$\pm$8.67\% & 0.974$\pm$0.64\% \\
				Log-Model & 0.113$\pm$30.69\% & 0.221$\pm$19.75\% & 0.616$\pm$3.44\% \\
				MUGL-\textit{o} & 0.485$\pm$12.65\% & 0.651$\pm$8.83\% & 0.964$\pm$0.67\% \\
				MUGL-\textit{l} & \textbf{0.490$\pm$11.93\%} & \textbf{0.657$\pm$8.36\%} & \textbf{0.964$\pm$0.65\%} \\
				\hline
			\end{tabular}
			\caption{COIL-20}
			\label{tab-coil20-sc}
		\end{subtable}
		\caption{Spectral clustering performance on image datasets}
		\label{tab-sc}
	\end{table}

\section{Conclusion}\label{sec-conclusion}
We have developed a novel DRO-based approach to graph learning, which provides a way to identify a graph that not only yields a smooth representation of the observed signals but is also robust against uncertainties about the ground-truth distribution of the graph signal. We have demonstrated how to construct the ambiguity set in our distributionally robust graph learning model by exploiting the structure of the Laplacian quadratic form and establishing confidence regions for the mean and covariance of the ground-truth distribution. We have also shown that whenever the ground-truth distribution has a probability density function, our proposed model admits a smooth non-convex optimization formulation. Interestingly, such a formulation provides a new perspective on regularization in the graph learning setting. Then, we have presented a PGD method to numerically tackle the formulation and established its convergence guarantees. Through extensive numerical experiments, we have shown that our proposed model improves the quality of the learned graphs and robustifies the performance across different populations of observed signals. One promising future direction is to extend our proposed approach to tackle more general graph learning scenarios.

\appendix
\subsection{Proof of Theorem \ref{thm-rho2}} \label{app:cr-cov}
To set the stage, let us introduce some additional notation. Given a real number $p\ge1$ and a $q_1\times q_2$ matrix $\bm{A}$, we use $\|\bm{A}\|_{S_p}$ to denote the Schatten $p$-norm of $\bm{A}$; i.e., $\|\bm{A}\|_{S_p} \coloneqq \|\sigma(\bm{A})\|_p$, where $\sigma(\bm{A}) \in \R_+^{\min\{q_1,q_2\}}$ is the vector of singular values of $\bm{A}$ and $\|\cdot\|_p$ is the usual vector $p$-norm. By definition, we have $\|\bm{A}\|_{S_2} = \|\bm{A}\|_F$.

We begin by establishing a relationship between the matrix
\[ \widetilde{\bm\Sigma}_n \coloneqq \frac{1}{n} \sum_{j=1}^n (\bm{x}^j - \bm{\mu}^*)(\bm{x}^j - \bm{\mu}^*)^\top \]
and the covariance matrix $\bm{\Sigma}^*$ of the ground-truth distribution $\bbP^*$. Note that the matrix $\widetilde{\bm\Sigma}_n$ is not the same as the empirical covariance matrix $\widehat{\bm\Sigma}_n$ defined in~\eqref{eq:sample-stat}, as the former is defined using $\bm{\mu}^*$ and not $\widehat{\bm\mu}_n$. Nevertheless, as we shall see, we can use the relationship between $\widetilde{\bm\Sigma}_n$ and $\bm{\Sigma}^*$ to establish the desired relationship between $\widehat{\bm\Sigma}_n$ and $\bm{\Sigma}^*$.
\begin{proposition} \label{prop:pseudo-cvar}
Under the setting of Theorem~\ref{thm-rho2}, there exists a constant $c_1>0$ such that 
\[ \| \widetilde{\bm\Sigma}_n - \bm{\Sigma}^* \|_F \le  \frac{4c_1(2e/3)^{3/2}\ln^{3/2}(4m^{3/2}/\delta)}{n^{1/2}} \|\bm{\Sigma}^*\| \]
will hold with probability at least $1-\delta/2$.
\end{proposition}
\begin{proof}
For $j=1,\ldots,n$, define
\[ 
\bm{Q}_j \coloneqq {{\bm\Sigma}^*}^{-1/2}(\bm{x}^j  - \bm{\mu}^*)(\bm{x}^j  - \bm{\mu}^*)^\top{{\bm\Sigma}^*}^{-1/2} - \bm{I}_m.
\] 
A straightforward calculation shows that $\E_{\bm{x}^j \sim \bbP^*}[ \bm{Q}_j ] = \bm{0}$ for $j=1,\ldots,n$. Moreover, since $\bbP^*$ satisfies the moment growth condition, there exists a constant $c' > 0$ such that for all $p\ge1$, 
\begin{equation} \label{eq:Sigma-mgc}
\E_{\bm{x} \sim \bbP^*} \left[ \| {\bm{\Sigma}^*}^{-1/2} ( \bm{x} - \bm{\mu} ) \|_2^p \right] \leq (c'p)^{p/2}.
\end{equation}
By combining~\eqref{eq:Sigma-mgc} with the argument in the proof of~\cite[Proposition 5]{so2011moment}, we deduce that for any $p\ge1$,
\[ \E_{\bm{x}^1,\ldots,\bm{x}^n \sim \bbP^*} \left[ \left\| \sum_{j=1}^n \bm{Q}_j \right\|_{S_p}^p \right] \le 2^p n^{p/2} p^{p/2} (m+(2c'p)^p). \]

Now, using the fact that $\|\bm{v}\|_2 \le q^{1/2} \|\bm{v}\|_{p}$ for any $\bm{v} \in \R^q$ and $p \in [2,+\infty]$ \footnote{The stated bound is not sharp but is sufficient for our purposes. Readers who are interested in the sharp bound can refer to, e.g.,~\cite[Lemma 1]{goldberg1987equivalence}.} and applying Markov's inequality, we have, for any $p\ge2$ and $t>0$, that
\begin{align*}
& \Pr\left( \left\| \frac{1}{n} \sum_{j=1}^n \bm{Q}_j \right\|_F > t \right) = \Pr\left( \left\| \frac{1}{n} \sum_{j=1}^n \bm{Q}_j \right\|_{S_2}^p > t^p \right) \\
&\le \frac{2^p p^{p/2} m^{1/2} (m+(2c'p)^p)}{t^p n^{p/2}}.
\end{align*}
In particular, by setting $c_1=\max\{c',1/4\}$, 
\[ t = \frac{4c_1(2e/3)^{3/2}\ln^{3/2}(4m^{3/2}/\delta)}{n^{1/2}}, \quad p = \left( \frac{tn^{1/2}}{4c_1e^{3/2}} \right)^{2/3} \]
and noting that $\delta \le e^{-2}$, we have $p=2\ln(4m^{3/2}/\delta)/3 \ge 2$ and
\[ 
\frac{2^p p^{p/2} m^{1/2} (m+(2c'p)^p)}{t^p n^{p/2}} = \frac{m^{3/2}+m^{1/2}(2c'p)^p}{e^{3p/2}(2c_1p)^p} \le \frac{\delta}{2}.
\] 
This, together with 
\[ \| \widetilde{\bm\Sigma}_n - \bm{\Sigma}^* \|_F = \left\| {\bm{\Sigma}^*}^{1/2} \left( \frac{1}{n} \sum_{j=1}^n \bm{Q}_j \right) {\bm{\Sigma}^*}^{1/2} \right\|_F, \]
implies the desired result.
\end{proof}

To proceed, observe that
\begin{align*}
& \widetilde{\bm\Sigma}_n = \frac{1}{n} \sum_{j=1}^n (\bm{x}^j - \widehat{\bm\mu}_n + \widehat{\bm\mu}_n - \bm{\mu}^*)(\bm{x}^j - \widehat{\bm\mu}_n + \widehat{\bm\mu}_n - \bm{\mu}^*)^\top \nonumber \\
&= \widehat{\bm\Sigma}_n + \frac{1}{n} \sum_{j=1}^n (\bm{x}^j - \widehat{\bm\mu}_n)(\widehat{\bm\mu}_n - \bm{\mu}^*)^\top \nonumber \\
&\quad+ \frac{1}{n}\sum_{j=1}^n (\widehat{\bm\mu}_n - \bm{\mu}^*) (\bm{x}^j - \widehat{\bm\mu}_n)^\top + (\widehat{\bm\mu}_n - \bm{\mu}^*) (\widehat{\bm\mu}_n - \bm{\mu}^*)^\top \nonumber \\
&= \widehat{\bm\Sigma}_n + (\widehat{\bm\mu}_n - \bm{\mu}^*) (\widehat{\bm\mu}_n - \bm{\mu}^*)^\top. \label{eq:pseudo-emp}
\end{align*}
Hence, we have
\begin{align*}
& \| \widehat{\bm\Sigma}_n - \bm{\Sigma}^* \|_F \le \| \widehat{\bm\Sigma}_n - \widetilde{\bm\Sigma}_n \|_F + \| \widetilde{\bm\Sigma}_n - \bm{\Sigma}^* \|_F \\
&=  (\widehat{\bm\mu}_n - \bm{\mu}^*)^\top (\widehat{\bm\mu}_n - \bm{\mu}^*) + \| \widetilde{\bm\Sigma}_n - \bm{\Sigma}^* \|_F.
\end{align*}
Since $\bbP^*$ satisfies the moment growth condition, by taking $c_2$ to be the constant $c$ in Definition~\ref{def:mgc} and adapting the proof of~\cite[Proposition 4]{so2011moment}, we deduce that with probability at least $1-\delta/2$, 
\[ (\widehat{\bm\mu}_n - \bm{\mu}^*)^\top (\widehat{\bm\mu}_n - \bm{\mu}^*) \le \frac{4c_2e^2\ln^2(2/\delta)}{n}. \]
This, together with Proposition~\ref{prop:pseudo-cvar}, implies that $\| \widehat{\bm\Sigma}_n - \bm{\Sigma}^* \|_F \le \widehat{\rho}_2$ will hold with probability at least $1-\delta$, as desired.

\vspace{-0.2cm}
\subsection{Proof of Proposition \ref{prop:closed-form}} \label{app:closed-form}
We first consider Problem~\eqref{eq:max-mu}. Upon letting $\widetilde{\bm\mu} = \bm{L}^{1/2}\bm{\mu}$, we can rewrite Problem~\eqref{eq:max-mu} as 
	\begin{equation}\label{eq:tilde}
		\begin{split}
			\sup_{\widetilde{\bm\mu}\in\mathbb{R}^m} &\ \|\widetilde{\bm\mu}\|_2^2\\
			\text{s.t.}\,\,\, & \,\, \|\widetilde{\bm\mu} - \bm{L}^{1/2} \widehat{\bm\mu}_n\|_2^2 \leq \rho_1^2.
		\end{split}
	\end{equation}
	Since the above problem satisfies the linear independence constraint qualification, its associated KKT conditions, which are given by
	\begin{equation*}
		\begin{split}
		-\widetilde{\bm\mu} + \lambda (\widetilde{\bm\mu} - \bm{L}^{1/2}\widehat{\bm\mu}_n) &= \bm{0}, \\
		\lambda \left(\|\widetilde{\bm\mu} - \bm{L}^{1/2}\widehat{\bm\mu}_n\|_2^2 - \rho_1^2\right) &=0,\\
		\lambda &\geq 0,\\
		\|\widetilde{\bm\mu} - \bm{L}^{1/2} \widehat{\bm\mu}_n\|_2^2 &\leq \rho_1^2,
		\end{split}
	\end{equation*}
are necessary for optimality. We consider the following two possibilities for the dual multiplier $\lambda$:

\smallskip	
\noindent\textit{Case I}: $\lambda=0$. The KKT conditions reduce to
\[ \widetilde{\bm\mu} = \bm{0}, \quad \| \bm{L}^{1/2} \widehat{\bm\mu}_n \|_2^2 \leq \rho_1^2. \]
The objective value of Problem~\eqref{eq:tilde} associated with the solution $\widetilde{\bm\mu} = \bm{0}$ is $0$, which is obviously not the maximum.

\smallskip
\noindent\textit{Case II}: $\lambda>0$. The KKT conditions become
\begin{subequations}
\begin{align}
-\widetilde{\bm\mu} + \lambda (\widetilde{\bm\mu} - \bm{L}^{1/2}\widehat{\bm\mu}_n) &= \bm{0}, \label{eq:II-1} \\
\|\widetilde{\bm\mu} - \bm{L}^{1/2}\widehat{\bm\mu}_n\|_2^2 &= \rho_1^2, \label{eq:II-2} \\
\lambda &> 0.
\end{align}
\end{subequations}
Let $\text{null}(\bm{L}^{1/2})$ denote the nullspace of $\bm{L}^{1/2}$. Consider the following two subcases:	
\begin{enumerate}
\item[(i)] If $\widehat{\bm\mu}_n \in \text{null}(\bm{L}^{1/2})$, then $\widetilde{\bm\mu}$ satisfies $\|\widetilde{\bm\mu}\|_2^2=\rho_1^2>0$ by~\eqref{eq:II-2}, which yields the objective value $\rho_1^2$.

\item[(ii)] If $\widehat{\bm\mu}_n \notin \text{null}(\bm{L}^{1/2})$, then $\lambda\neq1$. By~\eqref{eq:II-1}, we have $\widetilde{\bm\mu} = \tfrac{\lambda}{\lambda-1} \bm{L}^{1/2}\widehat{\bm\mu}_n$. Substituting this into~\eqref{eq:II-2} yields
\[ \lambda = 1\pm\frac{\|\bm{L}^{1/2}\widehat{\bm\mu}_n\|_2}{\rho_1}. \]
If $\lambda = 1 - \tfrac{\|\bm{L}^{1/2}\widehat{\bm\mu}_n\|_2}{\rho_1}$ and $\lambda>0$, then $\rho_1 > \| \bm{L}^{1/2}\widehat{\bm\mu}_n\|_2$ and
\[ \widetilde{\bm\mu} = \widetilde{\bm\mu}_- \coloneqq \left( 1 - \frac{\rho_1}{\|\bm{L}^{1/2}\widehat{\bm\mu}_n \|_2}\right) \bm{L}^{1/2} \widehat{\bm\mu}_n. \]
The objective value of Problem~\eqref{eq:tilde} associated with the solution $\widetilde{\bm\mu}_-$ is
\begin{align*}
\| \widetilde{\bm\mu}_- \|_2^2 &= \left( 1 - \frac{\rho_1}{\|\bm{L}^{1/2}\widehat{\bm\mu}_n \|_2}\right)^2 \| \bm{L}^{1/2} \widehat{\bm\mu}_n \|_2^2 \\
&= \left( \| \bm{L}^{1/2} \widehat{\bm\mu}_n \|_2 - \rho_1 \right)^2.
\end{align*}
On the other hand, if $\lambda = 1 + \tfrac{\|\bm{L}^{1/2}\widehat{\bm\mu}_n\|_2}{\rho_1}$, then 
\[ \widetilde{\bm\mu} = \widetilde{\bm\mu}_+ \coloneqq \left( 1 + \frac{\rho_1}{\|\bm{L}^{1/2}\widehat{\bm\mu}_n \|_2}\right) \bm{L}^{1/2} \widehat{\bm\mu}_n, \]
which yields the objective value 
\[ \| \widetilde{\bm\mu}_+ \|_2^2 = \left( \| \bm{L}^{1/2} \widehat{\bm\mu}_n \|_2 + \rho_1 \right)^2. \]
\end{enumerate}
Summarizing the above cases, we conclude that $\widetilde{\bm\mu}_+$ is an optimal solution to Problem~\eqref{eq:tilde}. This implies that $\varphi_1(\bm{L})=\left( \| \bm{L}^{1/2} \widehat{\bm\mu}_n \|_2 + \rho_1 \right)^2$, as desired.

Next, we consider Problem~\eqref{eq:max-sigma}. By dropping the constraint $\bm{\Sigma} \in \mathbb{S}_+^m$ from Problem~\eqref{eq:max-sigma}, we obtain the following relaxation:
\begin{equation}\label{eq-max-sigma-relax}
	\begin{split}
	\sup_{\bm{\Sigma}\in\mathbb{S}^m}&\ \tr(\bm{\Sigma} \bm{L}) \\
	\text{s.t.}\,\,\, & \,\,\,  \| \bm{\Sigma} - \widehat{\bm\Sigma}_n \|_F^2 \leq \rho_2^2.
	\end{split}
\end{equation}
Since the above problem is convex and satisfies the Slater condition, its associated KKT conditions, which are given by
\begin{subequations}
\begin{align}
-\bm{L} + 2\lambda(\bm{\Sigma}-\widehat{\bm\Sigma}_n) &= \bm{0}, \label{eq:Sig-1} \\
\| \bm{\Sigma} - \widehat{\bm\Sigma}_n \|_F^2 &\leq \rho_2^2, \\
\lambda \left(\|\bm{\Sigma}-\widehat{\bm\Sigma}_n\|_F^2-\rho_2^2\right) &= 0, \,\,\, \lambda \ge 0, \label{eq:Sig-2} 
\end{align}
\end{subequations}
are necessary and sufficient for optimality. Now, observe that we must have $\lambda>0$, for otherwise $\bm{L}=\bm{0}$ by~\eqref{eq:Sig-1}, which contradicts the fact that $\bm{L} \in \mathcal{L}_s$ satisfies $\tr(\bm{L})=2s>0$. Consequently, we have $\|\bm{\Sigma}-\widehat{\bm\Sigma}_n\|_F^2 = \rho_2^2$ by~\eqref{eq:Sig-2}. This, together with~\eqref{eq:Sig-1}, implies that
\[ \bm{\Sigma} = \bm{\Sigma}^* \coloneqq \widehat{\bm\Sigma}_n + \frac{\rho_2}{\|\bm{L}\|_F} \bm{L} \]
is an optimal solution to Problem~\eqref{eq-max-sigma-relax}. Since $\widehat{\bm\Sigma}_n, \bm{L} \in \mathbb{S}_+^m$ and $\rho_2>0$, we have $\bm{\Sigma}^*\in \mathbb{S}_+^m$. It follows that $\bm{\Sigma}^*$ is also optimal for Problem~\eqref{eq:max-sigma} and
\[
	\varphi_2(\bm{L})=\tr(\bm{\Sigma}^* \bm{L}) = \tr(\widehat{\bm\Sigma}_n \bm{L}) + \rho_2 \|\bm{L}\|_F.
\]

\bibliographystyle{IEEEtran}
\bibliography{ieeetrans}

\end{document}